\documentclass[conference]{IEEEtran}
\IEEEoverridecommandlockouts

\usepackage{cite}
\usepackage{amsmath,amssymb,amsfonts}
\usepackage{graphicx}
\usepackage{textcomp}
\usepackage{xcolor}
\usepackage{booktabs}
\usepackage{multirow}
\usepackage{url}
\usepackage{float}
\usepackage{subcaption}
\usepackage{hyperref}
\usepackage{balance}
\usepackage{array}
\usepackage{setspace}
\usepackage[explicit]{titlesec}
\usepackage[disable]{microtype}
\usepackage{fontspec}

\hypersetup{
    colorlinks=true,
    linkcolor=blue,
    citecolor=blue,
    urlcolor=blue
}

\newfontfamily\urdufont[
  Path = ./,
  Extension = .ttf,
  UprightFont = *-Regular,
  Script=Arabic,
  Scale=1.2
]{NotoNastaliqUrdu}

\newcommand{\urdText}[1]{\mbox{\urdufont\beginR#1\endR}}

\renewcommand{\normalsize}{\fontsize{10}{13}\selectfont}

\def\BibTeX{{\rm B\kern-.05em{\sc i\kern-.025em b}\kern-.08em
    T\kern-.1667em\lower.7ex\hbox{E}\kern-.125emX}}

\titleformat{\section}{\normalfont\large\bfseries\centering}{{\thesection}}{0.5em}{\MakeUppercase{#1}}
\titleformat{\subsection}{\normalfont\normalsize\bfseries}{{\thesubsection}}{0.5em}{#1}
\titleformat{\subsubsection}{\normalfont\normalsize\bfseries\itshape}{{\thesubsubsection}}{0.5em}{#1}
\titlespacing*{\section}{0pt}{1ex plus 0.5ex minus 0.2ex}{0.5ex plus 0.2ex}
\titlespacing*{\subsection}{0pt}{0.8ex plus 0.3ex minus 0.2ex}{0.3ex plus 0.1ex}
\titlespacing*{\subsubsection}{0pt}{0.6ex plus 0.3ex minus 0.1ex}{0.2ex plus 0.1ex}

\begin{document}

\title{When Benchmarks Mislead: Shortcut Learning, Length Confounds, and\\
the Limits of Cross-Dataset Generalization in Multilingual\\
Fake News and Sarcasm Detection}

\author{
\IEEEauthorblockN{Muhammad Abdullah Haroon}
\IEEEauthorblockA{
\textit{Department of Computer Science}\\
\textit{National University of Computer and Emerging Sciences (FAST-NUCES)}\\
Lahore, Pakistan\\
\href{mailto:abdullaharoon.work@gmail.com}{abdullaharoon.work@gmail.com}
}
}

\maketitle

\begin{abstract}
Cross-dataset generalisation is a fundamental requirement for deploying
text classifiers in real-world settings, yet systematic evaluation across
corpora from different sources remains uncommon in fake news detection
and virtually absent in sarcasm detection research. This paper presents
a unified empirical study of zero-shot cross-dataset transfer in three
domains: \textbf{Urdu fake news detection} (FND), \textbf{English FND},
and \textbf{sarcasm detection}. For each domain, we fine-tune
\texttt{xlm-roberta-base}~\cite{conneau2020unsupervised} on one corpus
and evaluate it on a second corpus from a different source, comparing
against TF-IDF baselines with Logistic Regression (LR) and Support
Vector Machines (SVM). In Urdu FND (Ax-to-Grind vs.\ Notri-Fact), we
identify a severe length confound in the Ax-to-Grind dataset
-- fake articles average 3.4$\times$ more words than real articles --
causing catastrophic A$\rightarrow$B transfer collapse (macro F1~=
0.005) while B$\rightarrow$A achieves F1~= 0.771. Extension to English
FND (WELFake vs.\ ISOT) and sarcasm detection (TweetEval Irony
vs.\ Sarcasm Corpus V2) reveals that such failure modes extend beyond
Urdu, confirming that shortcut learning from distributional artefacts
is a cross-lingual, cross-domain challenge in binary text classification.
We provide a reusable diagnostic methodology -- combining
class-conditional length analysis, bidirectional transfer asymmetry, and
predicted label collapse inspection -- applicable across any binary text
classification setting.
\end{abstract}


%%====================================================================
\section{Introduction}
\label{sec:intro}
%%====================================================================

The capacity of a trained model to perform well outside its training
distribution is a necessary -- but routinely untested -- property for
any practical NLP system. In fake news detection and sarcasm
detection, the standard evaluation protocol involves a single-dataset
80/20 train-test split~\cite{zhou2020survey,amjad2020bend}, which
masks shortcut learning: if a spurious surface feature (e.g., article
length, writing register) is equally present in both splits, the model
exploits it without penalty until deployed on data from a different
source.

This paper addresses that gap across three distinct classification
problems. The core case study is \textbf{Urdu fake news
detection}~(FND), where low-resource constraints amplify the risk that
models latch onto distributional artefacts. We demonstrate that a
systematic \textbf{length confound} in the Ax-to-Grind
corpus~\cite{harris2024axtogrind} -- fake articles average
3.4$\times$ more words than real articles -- causes a fine-tuned
XLM-RoBERTa to predict \emph{fake} for 99.7\% of a balanced
unseen corpus, collapsing macro F1 to 0.005. The reverse transfer
direction achieves F1~= 0.771, revealing the directional
asymmetry that exposure to a length-confounded corpus introduces.

To assess the generality of this finding, we extend the study to
\textbf{English FND} (WELFake~\cite{verma2021welfake} and the
ISOT dataset~\cite{ahmed2018isot}) and \textbf{sarcasm detection}
(TweetEval Irony~\cite{vanhee2018semeval,barbieri2020tweeteval}
and Sarcasm Corpus V2~\cite{oraby2016sarcasm}). These domains are
selected because they are (a) binary text classification tasks with
well-known corpora from structurally different sources, and (b) prone
to dataset-level confounds such as platform-specific writing style and
article length.

Our research questions are:
\begin{enumerate}
    \item \textbf{RQ1:} How well does XLM-RoBERTa generalise
    zero-shot across corpora in Urdu FND, English FND, and
    sarcasm detection?
    \item \textbf{RQ2:} When cross-domain transfer fails, what
    dataset-level properties explain the failure, and can the Urdu
    length-confound diagnostic methodology detect analogous failure
    modes in other domains?
    \item \textbf{RQ3:} Are length confounds and shortcut learning
    universal phenomena in these classification tasks, or are they
    specific to Urdu FND?
\end{enumerate}

\noindent Contributions:
\begin{itemize}
    \item The \textbf{first bidirectional cross-dataset transfer study}
    for Urdu FND, demonstrating catastrophic length-confound-driven
    collapse in one transfer direction and meaningful transfer in the
    reverse.
    \item \textbf{Empirical evidence of a severe length confound}
    in the Ax-to-Grind dataset (fake articles 3.4$\times$ longer)
    and quantification via length ablation (Exp.~5).
    \item \textbf{Extension to English FND} (WELFake
    $\leftrightarrow$ ISOT) and \textbf{sarcasm detection}
    (TweetEval $\leftrightarrow$ Sarcasm Corpus V2), enabling
    cross-domain comparison of generalisation behaviour.
    \item A \textbf{reusable diagnostic methodology} for identifying
    confound-driven failure in binary text classifiers.
    \item \textbf{TF-IDF baselines} (LR and SVM) across all six
    datasets, contextualising transformer performance.
\end{itemize}

%%====================================================================
\section{Related Work}
\label{sec:related}
%%====================================================================

\subsection{Urdu Fake News Detection}

Urdu FND is a nascent but growing research area. The earliest notable
contribution was the ``Bend the Truth'' dataset~\cite{amjad2020bend},
which formed the basis for the UrduFake@FIRE 2020 and 2021 shared
tasks~\cite{amjad2020fire,amjad2022fire}. Farooq et
al.~\cite{farooq2023fake} constructed a 4,097-article multi-domain
Urdu corpus using an ensemble of classical classifiers. Harris et
al.~\cite{harris2024axtogrind} introduced Ax-to-Grind Urdu (10,083
articles, 15 domains, Cohen's Kappa = 0.94), achieving F1~= 0.924
with a transformer ensemble; their analysis is exclusively in-domain.
Akhter et al.~\cite{akhter2021supervised} applied supervised ensemble
methods to Urdu social media FND.

\subsection{English Fake News Detection}

English FND has benefited from several large benchmark corpora. The
ISOT dataset~\cite{ahmed2018isot} pairs Reuters real news with articles
from flagged misinformation sites. The WELFake
dataset~\cite{verma2021welfake}, aggregated from four sources, has
become a standard evaluation benchmark. Silva et
al.~\cite{silva2021embracing} demonstrate that cross-domain transfer
within English FND frequently fails and propose multi-modal remedies.
Faustini and Covoes~\cite{faustini2020fake} study fake news detection
across multiple platforms and languages.

\subsection{Sarcasm and Irony Detection}

Sarcasm and irony detection present a closely related but
linguistically distinct classification challenge. Aboulenein et
al.~\cite{aboulenein2022sarcasm} survey a decade of progress,
identifying domain shift as a persistent obstacle. The TweetEval
Irony subset~\cite{barbieri2020tweeteval,vanhee2018semeval} provides a
Twitter-based binary irony classification benchmark. The Sarcasm
Corpus V2~\cite{oraby2016sarcasm}, derived from the Internet Argument
Corpus (IAC), covers three sub-categories: general sarcasm, hyperbole,
and rhetorical questions. The structural contrast between short,
informal Twitter text and longer forum posts makes these two corpora a
natural pair for cross-dataset transfer evaluation.

\subsection{Cross-Domain Fake News Detection}

Cross-domain generalisation in FND has been studied primarily for
English. Silva et al.~\cite{silva2021embracing} show that
domain-specific models fail under cross-domain evaluation and advocate
multi-modal robustness. Popat et al.~\cite{popat2018declare}
demonstrate that evidence-aware models transfer better than
content-only baselines. Zhou et al.~\cite{zhou2023fake} find that
linguistic surface features differ substantially across languages. Our
work provides the first cross-dataset study for Urdu FND and the first
comparative analysis spanning Urdu, English, and sarcasm detection.

\subsection{Multilingual Transformers}

XLM-RoBERTa~\cite{conneau2020unsupervised} is pre-trained on 2.5TB
of filtered CommonCrawl text across 100 languages including Urdu,
achieving strong results on XNLI, XQuAD, and MLQA.
mBERT~\cite{devlin2018bert} covers 104 languages. We use
\texttt{xlm-roberta-base} for all experiments to ensure comparability
across domains.

\subsection{Shortcut Learning and Dataset Artifacts}

Shortcut learning occurs when models exploit statistical correlations
between surface features and labels that do not generalise beyond the
training distribution. Geirhos et al.~\cite{geirhos2020shortcut}
provide a comprehensive characterisation of this failure mode across
deep network architectures, showing that models trained to minimise
empirical risk routinely exploit simple, non-semantic cues when these
are sufficiently predictive in training data. Wang and
Culotta~\cite{wang2021spurious} demonstrate the phenomenon in text
classification specifically, showing that classifiers exploit lexical
and stylistic surface correlations that degrade generalisation when
those correlations are absent at test time. Related evidence in
natural language inference includes Gururangan et
al.~\cite{gururangan2018annotation} (annotation artefacts) and McCoy
et al.~\cite{mccoy2019right} (syntactic heuristics). Our work
identifies an analogous \textbf{length-based shortcut} in fake news
and sarcasm detection -- tasks structurally distinct from NLI -- and
provides the first empirical evidence of this mechanism in Urdu NLP,
with comparative assessment across English and sarcasm datasets.

%%====================================================================
\section{Datasets}
\label{sec:datasets}
%%====================================================================

We use six datasets across three classification domains. Dataset
selection criteria: balanced class distributions, publicly available,
and structurally different sources within each domain pair (to
maximise the informational value of cross-dataset transfer).

\subsection{Urdu FND: Dataset A (Ax-to-Grind)}

The Ax-to-Grind corpus~\cite{harris2024axtogrind} contains 10,083
news articles across 15 domains. Articles were scraped from Urdu
newspaper and news channel websites (2017--2023); fake news was
gathered from fact-checking portals and social media captions. Expert
journalists verified every article (Cohen's Kappa~= 0.94). The
dataset is balanced: 5,030 real and 5,053 fake.

A structurally critical characteristic is its \textbf{length confound}.
As shown in Table~\ref{tab:dataset_stats}, fake articles average
116.98 words while real articles average only 34.82 words -- a
3.4$\times$ asymmetry. This arises from collection methodology:
genuine breaking news tends to be brief wire dispatches, while
fabricated narratives from misinformation websites tend to be longer
opinion pieces. This asymmetry creates a powerful spurious signal
available to any learned classifier.

\subsection{Urdu FND: Dataset B (Notri-Fact)}

The Notri-Fact dataset~\cite{notrifact2024} contains 13,388 Urdu
news articles (6,711 fake, 6,677 real). Each article concatenates
headline with body text. In contrast to Dataset A, Notri-Fact exhibits
\textbf{no significant length confound}: both classes average
approximately 160--170 words, making it a length-balanced benchmark.

\subsection{English FND: Dataset C (WELFake)}

The WELFake dataset~\cite{verma2021welfake} is a balanced English
fake news benchmark aggregating articles from four sources: Kaggle,
McIntire, Reuters, and BuzzFeed. It contains 72,134 articles with
labels 0~= Fake and 1~= Real. After deduplication and removal of null
entries, the working dataset comprises 63,121 articles
(Fake: 34,791; Real: 28,330). Fake articles average 892.6 words
while real articles average 679.0 words -- a moderate but notable
length asymmetry. Token length statistics on
\texttt{xlm-roberta-base}: mean~= 1,127, 90th pct~= 2,070,
setting MAX\_LENGTH~= 256. Table~\ref{tab:dataset_stats} reports
per-class word count statistics.

\subsection{English FND: Dataset D (ISOT)}

The ISOT dataset~\cite{ahmed2018isot} was compiled by the
Information Security and Object Technology research lab at the
University of Victoria, pairing Reuters real news with articles from
websites flagged by PolitiFact and Wikipedia. The combined dataset
contains approximately 44,898 articles. For computational efficiency,
a stratified random sample of 20,000 articles (10,000 per class) was
drawn for all experiments, with each article represented as the
concatenation of its title and body text. Labels: 0~= Fake, 1~= Real.
Fake articles average 429.0 words and real articles average 394.8
words -- an approximately balanced length distribution.
Per-class word count statistics are reported in
Table~\ref{tab:dataset_stats}.

\subsection{Sarcasm Detection: Dataset E (TweetEval Irony)}

The TweetEval Irony subset~\cite{barbieri2020tweeteval,vanhee2018semeval}
corresponds to SemEval 2018 Task 3 (binary irony detection). Combined
train, validation, and test splits yield 4,601 tweets
(not ironic: 2,389; ironic: 2,212). Labels: 0~= not ironic,
1~= ironic. Both classes exhibit near-identical word count
distributions (mean~$\approx$~13.7 words), reflecting the uniform
brevity of Twitter content.

\subsection{Sarcasm Detection: Dataset F (Sarcasm Corpus V2)}

The Sarcasm Corpus V2~\cite{oraby2016sarcasm} is drawn from the
Internet Argument Corpus (IAC), comprising online debate forum posts
annotated for sarcasm by independent annotators across three
sub-categories: general sarcasm (GEN), hyperbole (HYP), and rhetorical
questions (RQ). All three sub-categories are concatenated into a single
dataset for our experiments, yielding 9,386 posts (not sarcastic:
4,693; sarcastic: 4,693). Labels: 0~= not sarcastic
(\texttt{notsarc}), 1~= sarcastic (\texttt{sarc}). Not-sarcastic
posts average 55.8 words while sarcastic posts average 41.6 words
(Table~\ref{tab:dataset_stats}), creating a moderate within-corpus
length difference. Crucially, all Sarcasm Corpus V2 posts are
substantially longer than TweetEval tweets (13.7 words), creating a
pronounced cross-corpus length gap that mirrors, in structure, the
confound found in Dataset A.

\subsection{Schema Alignment}

All datasets were standardised to a common two-column schema:
\texttt{text} and \texttt{label} (0~= negative class, 1~= positive
class). For Dataset A: \texttt{News Items} $\to$ \texttt{text}; FAKE
$\to$ 0, TRUE $\to$ 1. For Dataset B: headline and body concatenated
into \texttt{text}; Unreal $\to$ 0, Real $\to$ 1. For Datasets C/D:
\texttt{title} + \texttt{text} concatenated into \texttt{news};
conventions preserved as above. For Dataset E: native HuggingFace
\texttt{text}/\texttt{label} columns used directly. For Dataset F:
\texttt{class} renamed to \texttt{label}; \texttt{notsarc} $\to$ 0,
\texttt{sarc} $\to$ 1. Zero overlap between corpora in each domain
pair was verified by exact string deduplication.

\begin{table*}[!htb]
\centering
\caption{Dataset Statistics and Per-Class Length Analysis Across All Six Datasets.
$^{\dagger}$Approximate; $^{\ddagger}$Sampled to 20,000 for computational efficiency.}
\label{tab:dataset_stats}
\setlength{\tabcolsep}{4pt}
\begin{tabular}{llllcc}
\toprule
\textbf{Domain} & \textbf{Dataset} & \textbf{Label} & \textbf{Class} &
\textbf{Count} & \textbf{Avg Words} \\
\midrule
\multirow{6}{*}{Urdu FND}
  & \multirow{3}{*}{A: Ax-to-Grind}
    & Real  & 1 & 5,030  & 34.82 \\
  & & Fake  & 0 & 5,053  & 116.98 \\
  & & Total &   & 10,083 & 75.90 \\
\cmidrule{2-6}
  & \multirow{3}{*}{B: Notri-Fact}
    & Real  & 1 & 6,677  & $\approx$160 \\
  & & Fake  & 0 & 6,711  & $\approx$170 \\
  & & Total &   & 13,388 & $\approx$165 \\
\midrule
\multirow{6}{*}{English FND}
  & \multirow{3}{*}{C: WELFake}
    & Real  & 1 & 28,330 & 679.0 \\
  & & Fake  & 0 & 34,791 & 892.6 \\
  & & Total &   & 63,121 & 769.1 \\
\cmidrule{2-6}
  & \multirow{3}{*}{D: ISOT$^{\ddagger}$}
    & Real  & 1 & 10,000 & 394.8 \\
  & & Fake  & 0 & 10,000 & 429.0 \\
  & & Total &   & 20,000 & 411.9 \\
\midrule
\multirow{6}{*}{Sarcasm}
  & \multirow{3}{*}{E: TweetEval}
    & Not ironic & 0 & 2,389 & 13.7 \\
  & & Ironic     & 1 & 2,212 & 13.8 \\
  & & Total      &   & 4,601 & 13.7 \\
\cmidrule{2-6}
  & \multirow{3}{*}{F: Sarcasm Corpus V2}
    & Not sarc   & 0 & 4,693 & 55.8 \\
  & & Sarcastic  & 1 & 4,693 & 41.6 \\
  & & Total      &   & 9,386 & 48.7 \\
\bottomrule
\end{tabular}
\end{table*}

\subsection{Vocabulary and Word Cloud Analysis -- Urdu}
\label{sec:vocab}

\begin{figure*}[!htb]
    \centering
    \IfFileExists{top20DatasetA.png}{\includegraphics[width=\textwidth]{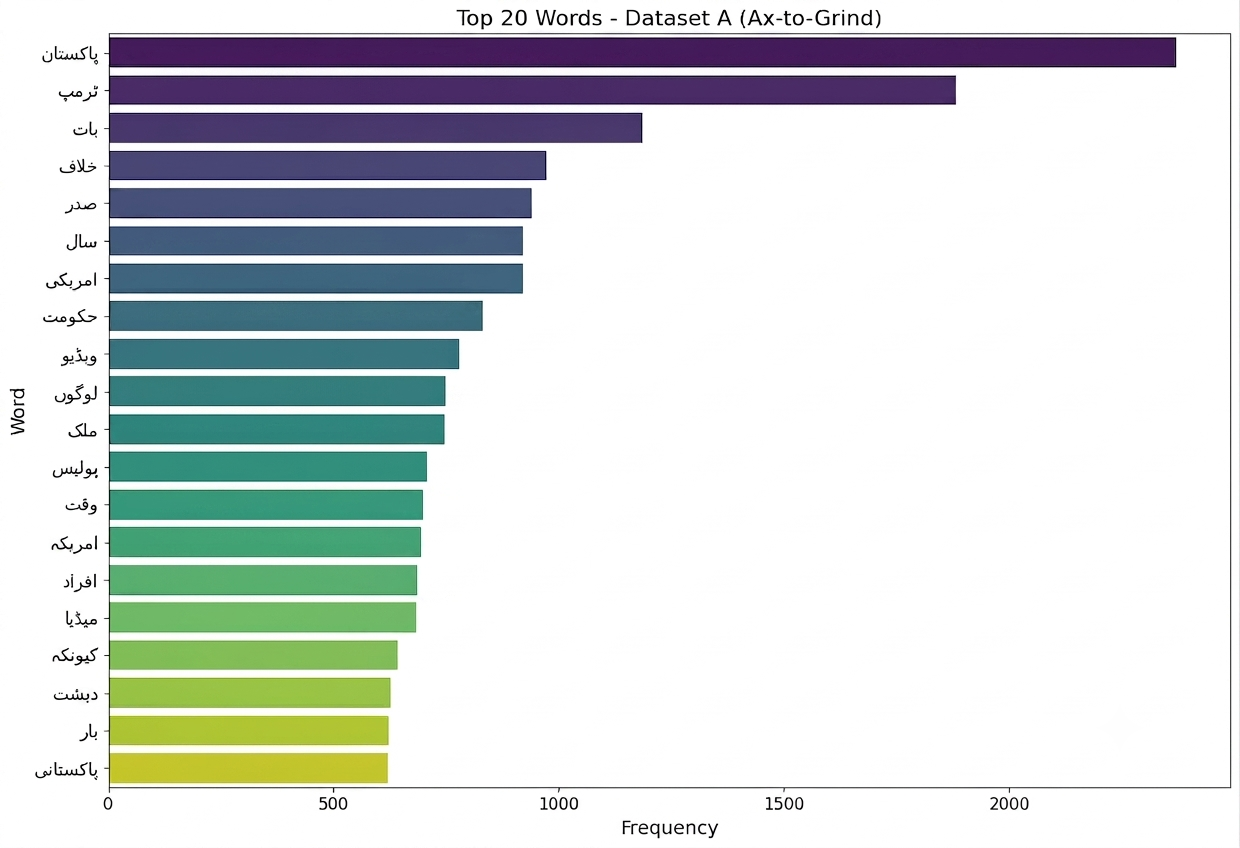}}{\fbox{\parbox{\textwidth}{\centering\vspace{2cm}\textit{[Figure: Top-20 words -- Dataset A -- image file top20DatasetA.png not yet uploaded]}\vspace{2cm}}}}
    \caption{%
      \textbf{Dataset A (Ax-to-Grind) -- Top-20 Most Frequent Content Words}
      after Urdu grammatical stopword removal. Bar length indicates total
      occurrences. Geopolitical and political terms dominate: Pakistan,
      Trump, ``against,'' president. The political skew partially explains
      B$\rightarrow$A transfer success, as Notri-Fact shares this
      geopolitical lexicon.}
    \label{fig:freqA}
\end{figure*}

\begin{figure*}[!htb]
    \centering
    \IfFileExists{top20DatasetB.png}{\includegraphics[width=\textwidth]{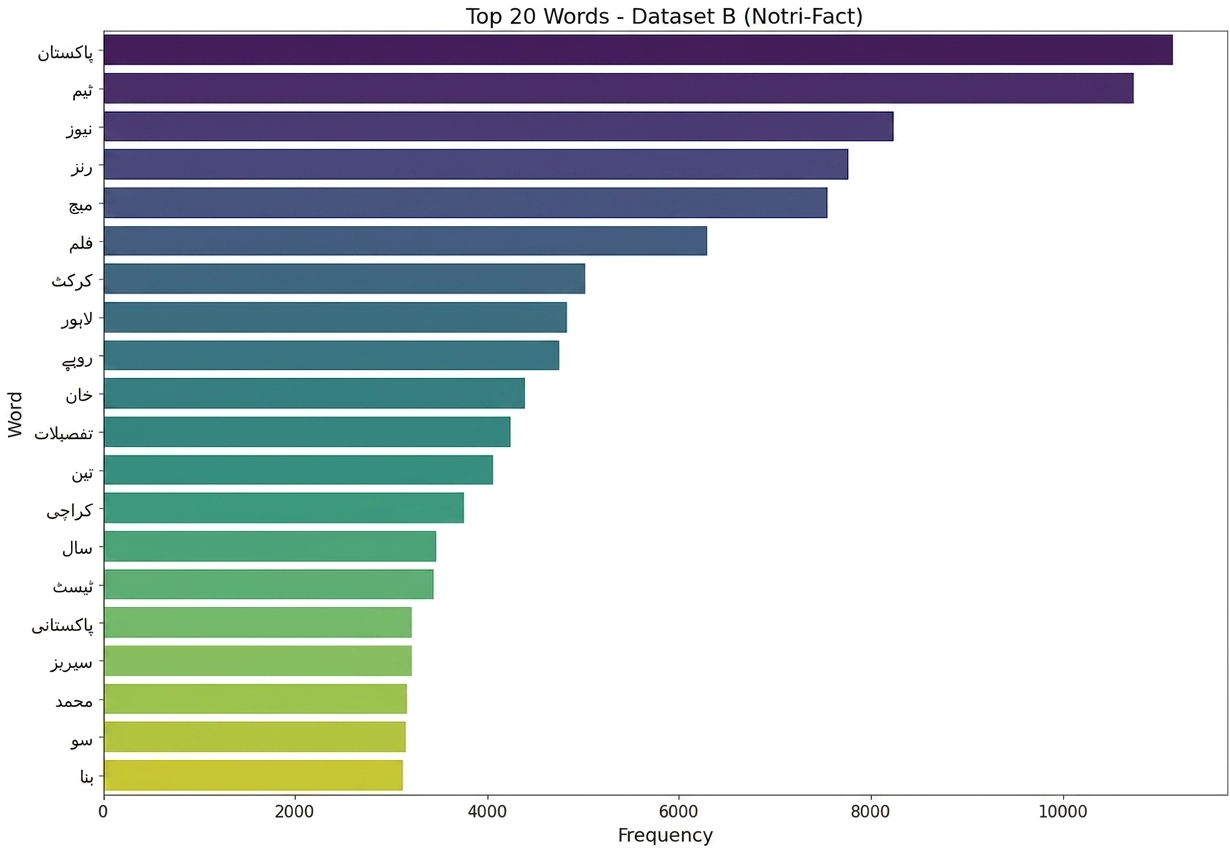}}{\fbox{\parbox{\textwidth}{\centering\vspace{2cm}\textit{[Figure: Top-20 words -- Dataset B -- image file top20DatasetB.png not yet uploaded]}\vspace{2cm}}}}
    \caption{%
      \textbf{Dataset B (Notri-Fact) -- Top-20 Most Frequent Content Words}
      after the same stopword preprocessing. Cricket and entertainment
      vocabulary signals a topically broader corpus. Higher frequency
      ceilings reflect Notri-Fact's larger size (13,388 vs.\ 10,083
      articles).}
    \label{fig:freqB}
\end{figure*}

Figures~\ref{fig:freqA} and~\ref{fig:freqB} present the top-20 most frequent content words for each Urdu dataset; these were
computed after applying a comprehensive Urdu stopword list. Removed
tokens include common particles such as
\urdText{کا}~(of), \urdText{کی}~(of/from), \urdText{میں}~(in),
\urdText{ہے}~(is), \urdText{نے}~(subject marker), \urdText{سے}~(from),
\urdText{اور}~(and), \urdText{نہیں}~(not), \urdText{لیکن}~(but),
along with Urdu punctuation marks
(\urdText{،}~comma, \urdText{۔}~full stop,
\urdText{؟}~question mark, \urdText{؛}~semicolon).

Dataset~A top words (\urdText{پاکستان}~Pakistan, \urdText{ٹرمپ}~Trump,
\urdText{خلاف}~against, \urdText{صدر}~president) reflect geopolitical
framing. Dataset~B top words (\urdText{ٹیم}~team, \urdText{رنز}~runs,
\urdText{میچ}~match, \urdText{فلم}~film) signal a broader topical scope.
The shared high-frequency term \urdText{پاکستان} provides lexical common
ground partially explaining the meaningful B$\rightarrow$A transfer
(F1~= 0.771). Word clouds for Dataset~A show the density and topical
contrast between fake and real articles.

\begin{figure*}[!htb]
    \centering
    \IfFileExists{Fake.png}{\includegraphics[width=\textwidth]{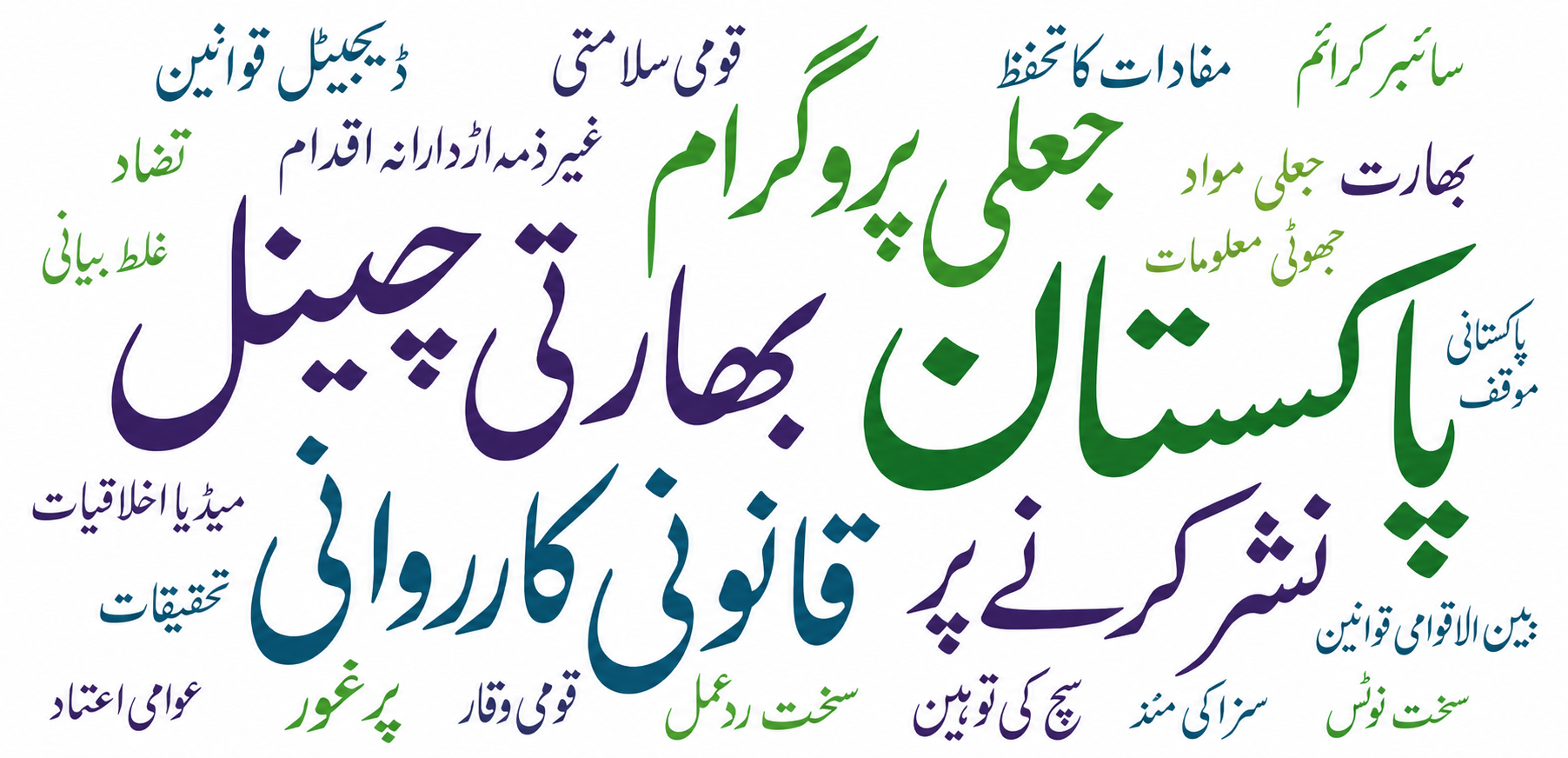}}{\fbox{\parbox{\textwidth}{\centering\vspace{2cm}\textit{[Figure: Word Cloud -- Fake articles -- image file Fake.png not yet uploaded]}\vspace{2cm}}}}
    \caption{%
      \textbf{Word Cloud -- Fake Articles, Dataset A (Ax-to-Grind).}
      Word size proportional to TF-IDF weight. Geopolitical and
      conspiratorial framing is prominent, consistent with politically
      motivated misinformation. High visual density reflects the longer
      average length of fake articles (117 words).}
    \label{fig:wcFake}
\end{figure*}

\begin{figure*}[!htb]
    \centering
    \IfFileExists{tn.jpg}{\includegraphics[width=\textwidth]{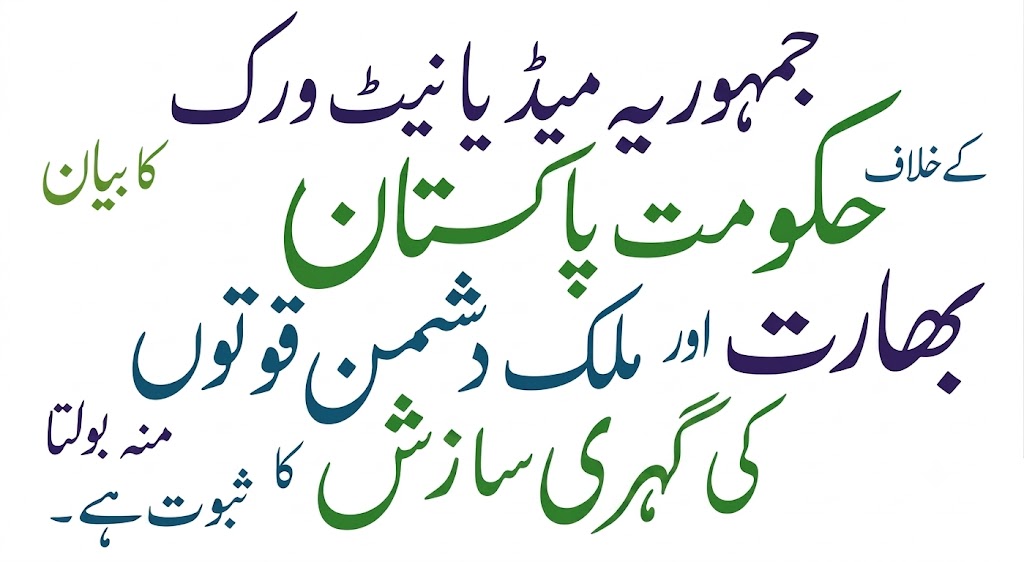}}{\fbox{\parbox{\textwidth}{\centering\vspace{2cm}\textit{[Figure: Word Cloud -- Real articles -- image file True.png not yet uploaded]}\vspace{2cm}}}}
    \caption{%
      \textbf{Word Cloud -- Real Articles, Dataset A (Ax-to-Grind).}
      Sparser and more factual vocabulary consistent with brief
      wire-dispatch reporting. Lower visual density reflects shorter
      average length of real articles (35 words), the structural basis
      of the length confound.}
    \label{fig:wcTrue}
\end{figure*}

\begin{figure*}[!htb]
    \centering
    \includegraphics[width=\textwidth]{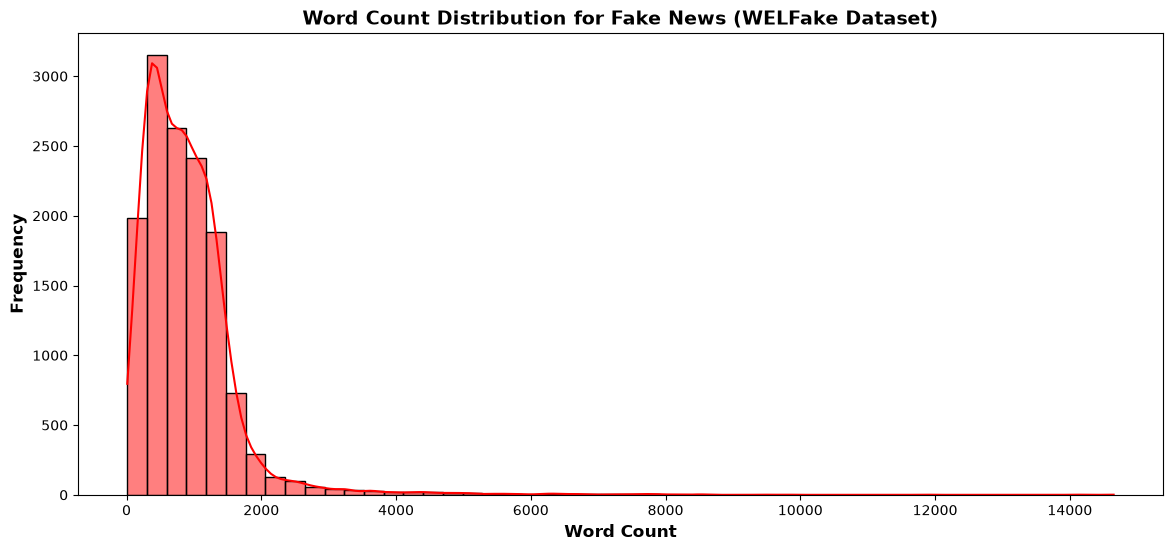}
    \caption{%
      \textbf{English FND -- WELFake (Dataset C): Fake Class Word Count Distribution.}
      Fake articles (label~0) average 892.6 words, peaking around
      300--400 words with a long right tail extending to 14,000+ words.}
    \label{fig:distEN_fake}
\end{figure*}

\begin{figure*}[!htb]
    \centering
    \includegraphics[width=\textwidth]{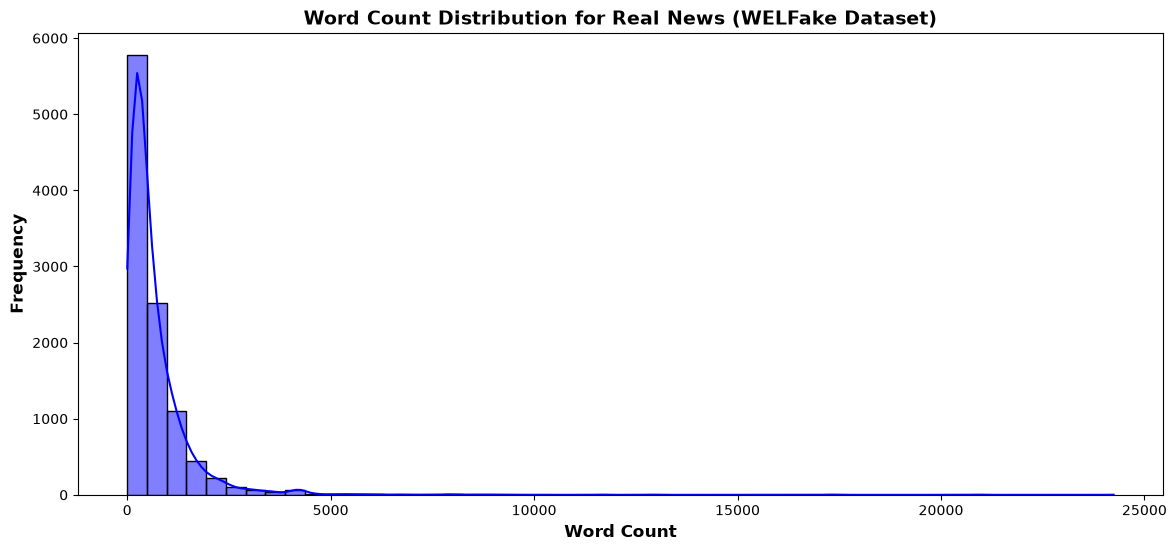}
    \caption{%
      \textbf{English FND -- WELFake (Dataset C): Real Class Word Count Distribution.}
      Real articles (label~1) average 679.0 words, with a sharper
      distribution peaking below 500 words. The moderate difference in
      class means (892 vs.\ 679 words) suggests a weaker length signal
      compared to Ax-to-Grind's 3.4$\times$ ratio.}
    \label{fig:distEN_real}
\end{figure*}

\begin{figure*}[!htb]
    \centering
    \includegraphics[width=\textwidth]{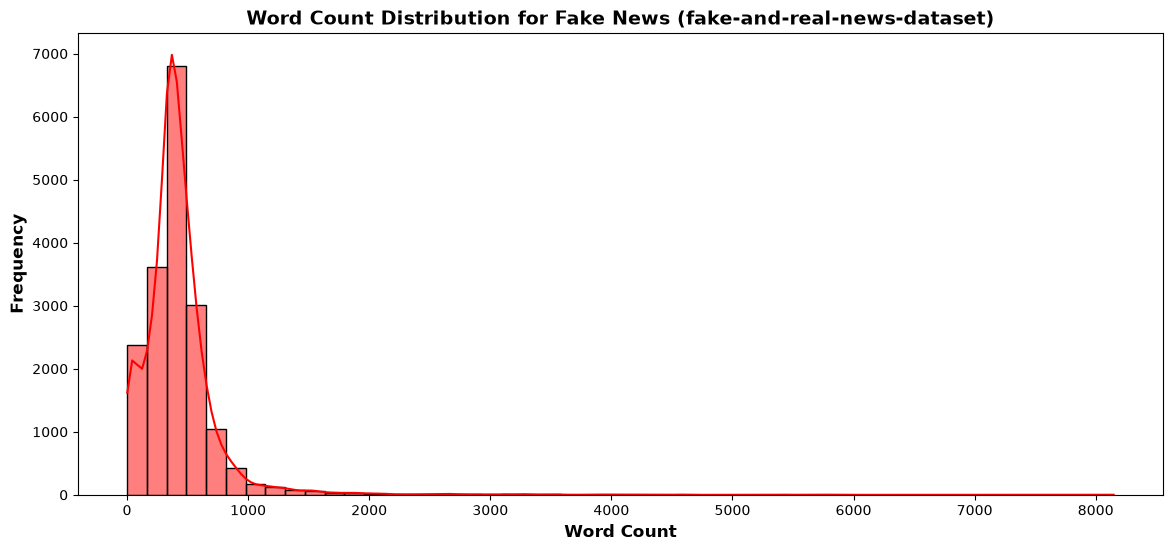}
    \caption{%
      \textbf{English FND -- ISOT (Dataset D): Fake Class Word Count Distribution.}
      Fake articles average 429.0 words, peaking around 300--600 words.}
    \label{fig:distEN_isot_fake}
\end{figure*}

\begin{figure*}[!htb]
    \centering
    \includegraphics[width=\textwidth]{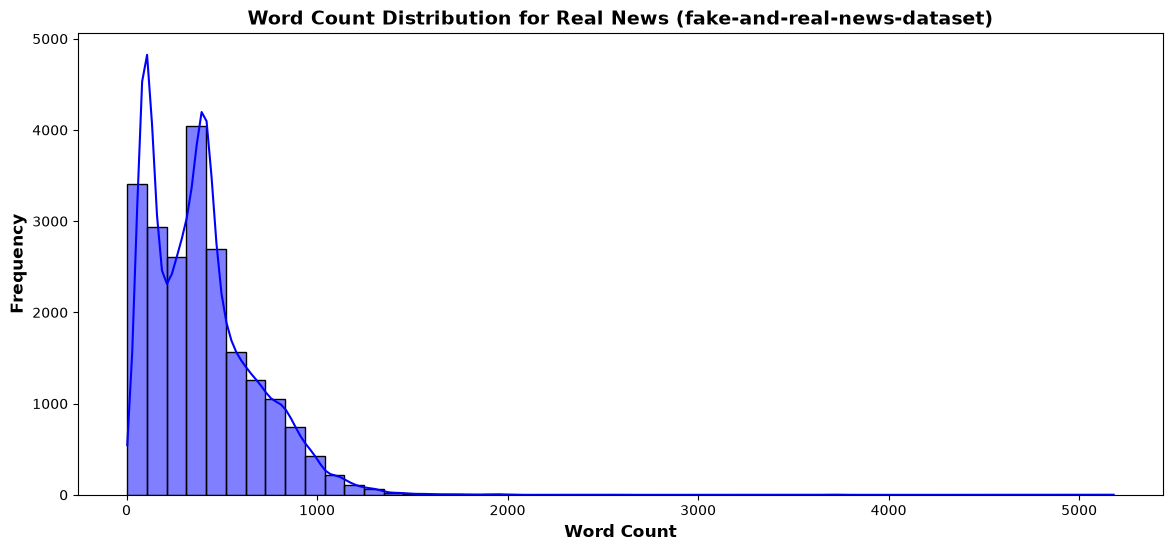}
    \caption{%
      \textbf{English FND -- ISOT (Dataset D): Real Class Word Count Distribution.}
      Real articles average 394.8 words, with a bimodal distribution
      reflecting the mix of short and long Reuters articles.
      The near-symmetric length distributions between fake (429 words)
      and real (395 words) ISOT articles suggest a minimal within-corpus
      length confound.}
    \label{fig:distEN_isot_real}
\end{figure*}

\begin{figure*}[!htb]
    \centering
    \includegraphics[width=\textwidth]{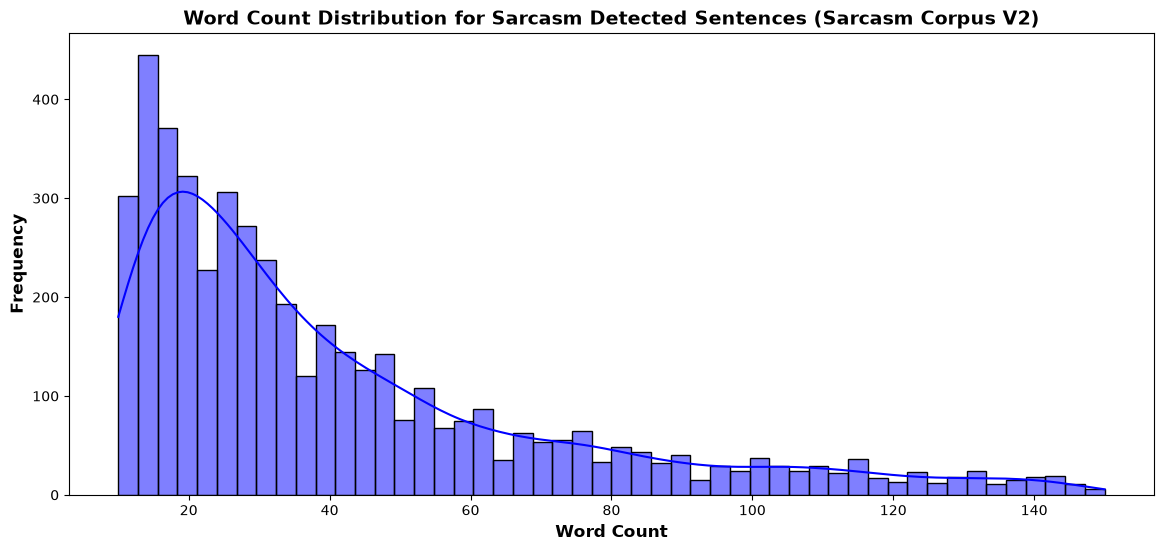}
    \caption{%
      \textbf{Sarcasm -- Sarcasm Corpus V2 (Dataset F): Sarcastic Class Word Count Distribution.}
      Sarcastic forum posts (label~1) average 41.6 words, right-skewed
      up to 150+ words.}
    \label{fig:distSarc_v2sar}
\end{figure*}

\begin{figure*}[!htb]
    \centering
    \includegraphics[width=\textwidth]{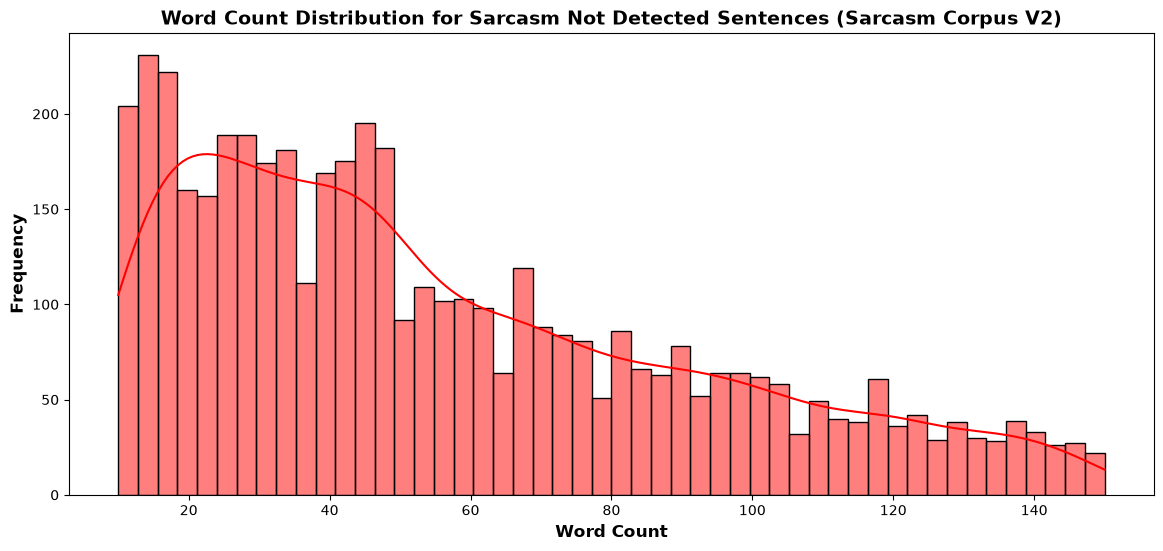}
    \caption{%
      \textbf{Sarcasm -- Sarcasm Corpus V2 (Dataset F): Non-Sarcastic Class Word Count Distribution.}
      Non-sarcastic forum posts (label~0) average 55.8 words with a
      flatter distribution, notably longer than sarcastic posts.}
    \label{fig:distSarc_v2nosar}
\end{figure*}

\begin{figure*}[!htb]
    \centering
    \includegraphics[width=\textwidth]{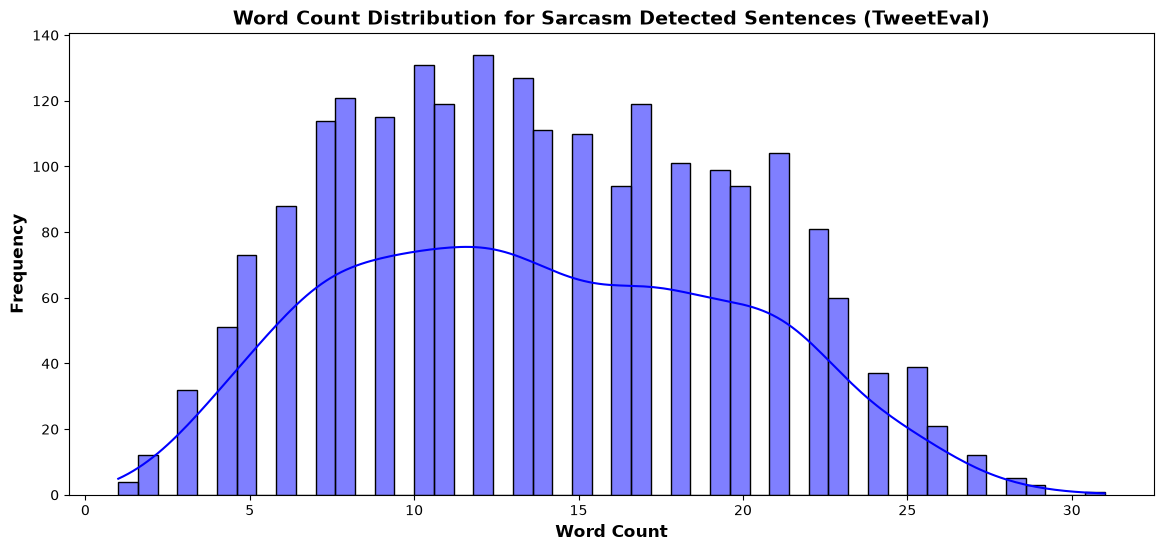}
    \caption{%
      \textbf{Sarcasm -- TweetEval Irony (Dataset E): Ironic Class Word Count Distribution.}
      Ironic tweets (label~1) average 13.8 words, concentrated between
      5--25 words, reflecting Twitter's character limit.}
    \label{fig:distSarc_tweetSar}
\end{figure*}

\begin{figure*}[!htb]
    \centering
    \includegraphics[width=\textwidth]{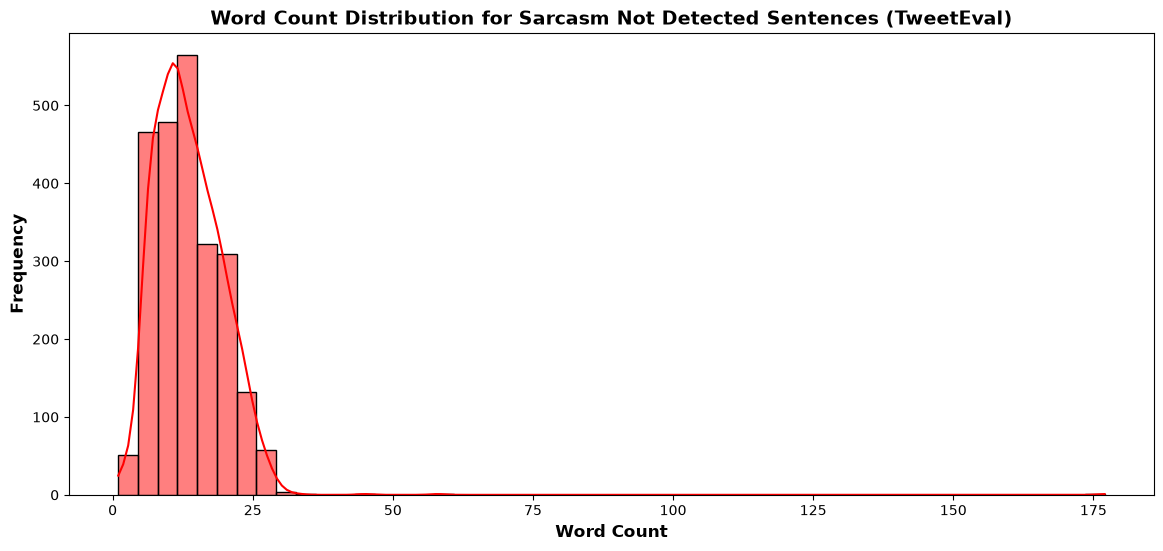}
    \caption{%
      \textbf{Sarcasm -- TweetEval Irony (Dataset E): Non-Ironic Class Word Count Distribution.}
      Non-ironic tweets (label~0) average 13.7 words, nearly identical
      in distribution to the ironic class. The near-zero within-corpus
      length difference stands in sharp contrast to the Sarcasm Corpus
      V2 (avg.\ 48.7 words), creating a substantial cross-corpus length
      gap that mirrors the Urdu case.}
    \label{fig:distSarc_tweetNoSar}
\end{figure*}

Figures~\ref{fig:distEN_fake}--\ref{fig:distSarc_tweetNoSar}
show the \textbf{text-length} (word count) distributions for the English
FND and sarcasm domain pairs.

%%-- English FND length distributions --%%

%%-- Sarcasm length distributions --%%

%%====================================================================
\section{Methodology}
\label{sec:methodology}
%%====================================================================

\subsection{XLM-RoBERTa Fine-Tuning}

We use \texttt{xlm-roberta-base}~\cite{conneau2020unsupervised} for
all six datasets and all transfer experiments. The model is fine-tuned
for binary sequence classification by attaching a linear classification
head to the \texttt{[CLS]} token representation.

\textbf{Preprocessing.} For XLM-RoBERTa, only whitespace stripping is
applied prior to tokenisation. Aggressive cleaning (stopword removal,
stemming) is deliberately avoided, as it degrades multilingual
transformer performance by shifting inputs away from the pre-training
distribution.

\textbf{Token length.} MAX\_LENGTH~= 256, covering $\approx$90\% of
all corpora. Percentile analysis on Dataset~A: mean~= 96, median~= 39,
90th~= 193, 95th~= 545, 99th~= 859 tokens. For Dataset~C
(WELFake): mean~= 1,127, 90th pct~= 2,070. For Dataset~D (ISOT):
mean~= 558, 90th pct~= 1,025. For Dataset~E (TweetEval): mean~= 25,
90th pct~= 38. For Dataset~F (Sarcasm Corpus V2): mean~= 70,
90th pct~= 141.

\textbf{Hyperparameters.} 3 epochs (Datasets A/B/C/E/F), 2 epochs
(Dataset~D, ISOT); learning rate $2\times10^{-5}$; batch size 16;
weight decay 0.01; AdamW optimiser; linear warmup schedule; seed 42
throughout. All experiments on NVIDIA T4 GPU via Google Colab.

\subsection{TF-IDF Baselines}

A language-appropriate cleaning pipeline is applied: Urdu punctuation
marks, English punctuation, digits, URLs, and domain-specific
stopwords are removed. For sarcasm Twitter data, URLs, emojis, and
HTML entities are additionally removed. Unigrams and bigrams are
extracted with sublinear TF weighting and minimum document frequency~2.
Two classifiers:
\begin{itemize}
    \item \textbf{Logistic Regression (LR):} L2 regularisation,
    $C$~= 1.0, liblinear solver, 1000 iterations.
    \item \textbf{SVM:} Linear kernel, hinge loss, $C$~= 1.0.
\end{itemize}

\subsection{Experimental Design}
\label{sec:expdesign}

Table~\ref{tab:experiments} details all 13 experiments.
Experiments~3, 4, 8, 9, 12, and 13 are inference-only, reusing
saved checkpoint weights from their respective in-domain training
runs. This constitutes strict \textbf{zero-shot cross-dataset
transfer} with no adaptation on the target corpus.

\begin{table*}[!htb]
\centering
\caption{Experimental Conditions}
\label{tab:experiments}
\setlength{\tabcolsep}{4pt}
\begin{tabular}{clll}
\toprule
\textbf{Exp} & \textbf{Name} & \textbf{Train} & \textbf{Test} \\
\midrule
\multicolumn{4}{l}{\textit{Urdu Fake News Detection}} \\
1  & In-domain A            & df\_a 80\%          & df\_a 20\%      \\
2  & In-domain B            & df\_b 80\%          & df\_b 20\%      \\
3  & Cross A$\to$B          & Full df\_a          & Full df\_b      \\
4  & Cross B$\to$A          & Full df\_b          & Full df\_a      \\
5  & Length Ablation A      & df\_a 80\% (50-word cap) & df\_a 20\% (50-word cap) \\
\midrule
\multicolumn{4}{l}{\textit{English Fake News Detection}} \\
6  & In-domain C            & df\_c 80\%          & df\_c 20\%      \\
7  & In-domain D            & df\_d 80\%          & df\_d 20\%      \\
8  & Cross C$\to$D          & Full df\_c          & Full df\_d      \\
9  & Cross D$\to$C          & Full df\_d          & Full df\_c      \\
\midrule
\multicolumn{4}{l}{\textit{Sarcasm Detection}} \\
10 & In-domain E            & df\_e 80\%          & df\_e 20\%      \\
11 & In-domain F            & df\_f 80\%          & df\_f 20\%      \\
12 & Cross E$\to$F          & Full df\_e          & Full df\_f      \\
13 & Cross F$\to$E          & Full df\_f          & Full df\_e      \\
\bottomrule
\end{tabular}
\end{table*}

All 80/20 splits are stratified by label. Experiment~5 (length
ablation) truncates all Dataset~A articles to 50 words before
training and evaluation, isolating the effect of the length confound
on in-domain performance.

\textbf{Evaluation.} Primary metric: \textbf{macro-averaged F1-score},
robust to class imbalance and standard for all three task types.
Accuracy, precision, and recall are reported where available.

\clearpage
%%====================================================================
\section{Results and Analysis}
\label{sec:results}
%%====================================================================

\subsection{Urdu FND Results}

Table~\ref{tab:main_results_urdu} presents complete results for
Experiments~1--5. XLM-RoBERTa achieves strong in-domain performance:
F1~= 0.929 on Dataset~A and F1~= 0.954 on Dataset~B. TF-IDF
baselines are consistently competitive but trail the transformer by
$\approx$7 percentage points on Dataset~A and $\approx$4 points on
Dataset~B.

\begin{table*}[!htb]
\centering
\caption{Full Results: Urdu Fake News Detection}
\label{tab:main_results_urdu}
\begin{tabular}{llcccc}
\toprule
\textbf{Exp} & \textbf{Model} & \textbf{F1} & \textbf{Acc} &
\textbf{Prec} & \textbf{Rec} \\
\midrule
\multicolumn{6}{l}{\textit{Dataset A -- Ax-to-Grind (in-domain, 20\% test split)}} \\
1 & XLM-RoBERTa              & 0.929 & 0.929 & 0.929 & 0.929 \\
1 & XLM-RoBERTa (50-word cap, Exp 5) & 0.922 & 0.922 & 0.922 & 0.922 \\
- & TF-IDF + LR              & 0.860 & 0.860 & 0.860 & 0.860 \\
- & TF-IDF + SVM             & 0.855 & 0.855 & 0.855 & 0.855 \\
\midrule
\multicolumn{6}{l}{\textit{Dataset B -- Notri-Fact (in-domain, 20\% test split)}} \\
2 & XLM-RoBERTa              & 0.954 & 0.953 & 0.954 & 0.954 \\
- & TF-IDF + LR              & 0.900 & 0.900 & 0.900 & 0.900 \\
- & TF-IDF + SVM             & 0.910 & 0.910 & 0.910 & 0.910 \\
\midrule
\multicolumn{6}{l}{\textit{Cross-Domain Transfer (zero-shot, full dataset)}} \\
3 & XLM-RoBERTa (A$\to$B)   & \textbf{0.005} & 0.501 & 0.250 & 0.501 \\
4 & XLM-RoBERTa (B$\to$A)   & 0.771 & 0.773 & 0.774 & 0.773 \\
\bottomrule
\end{tabular}
\end{table*}

\subsubsection{Cross-Domain Transfer: Asymmetric Collapse}

The cross-domain results reveal a dramatic asymmetry.
\textbf{Experiment~4 (B$\rightarrow$A):} F1~= 0.771 represents an
18.3-point drop from in-domain performance but confirms meaningful
transfer: the model retains substantial discriminative ability in the
unseen domain without any Ax-to-Grind training data.

\textbf{Experiment~3 (A$\rightarrow$B):} F1~= 0.005. Inspection of
the predicted label distribution (Table~\ref{tab:prediction_collapse})
reveals the failure mechanism: the model predicts ``fake'' (label~0)
for 13,350 out of 13,388 Notri-Fact articles (99.7\%), with only 38
predicted real. The actual distribution is 6,711 fake and 6,677 real
-- balanced. The model has collapsed to a near-constant prediction of
one class.

\begin{table}[h!]
\centering
\caption{Predicted Label Distribution: Experiment~3 (A$\to$B)}
\label{tab:prediction_collapse}
\begin{tabular}{lcc}
\toprule
 & \textbf{Pred.\ Fake} & \textbf{Pred.\ Real} \\
\midrule
Predicted count      & 13,350 & 38 \\
Actual count         & 6,711  & 6,677 \\
Proportion predicted & 99.7\% & 0.3\% \\
\bottomrule
\end{tabular}
\end{table}

\subsubsection{Diagnosing the Collapse: The Length Confound}

Three steps explain the A$\to$B collapse.

\textbf{Step 1 -- Shortcut availability.} During fine-tuning on
Dataset~A, the model has access to a trivially learnable
feature: token sequence length. Fake articles average 117 words
while real articles average 35 words. A classifier can achieve high
accuracy simply by learning that long sequences are fake.

\textbf{Step 2 -- Shortcut exploitation.} Transformers exploit the
easiest available feature rather than learning the intended semantic
task~\cite{geirhos2020shortcut, wang2021spurious}. The 3.4$\times$
length ratio likely causes the model's attention mechanism to encode
length-related signals -- effectively treating long documents as fake
regardless of content.

\textbf{Step 3 -- Distribution mismatch at test time.} Notri-Fact
articles average 160--170 words regardless of class. Every
Notri-Fact article looks like a ``fake'' article by the model's
learned heuristic. The result is near-constant fake prediction.

This hypothesis is supported by the \textbf{extreme} prediction skew.
A simple generalisation failure would yield near-chance F1~$\approx$~0.50;
the observed F1~$\approx$~0.005 is only possible when the model has
learned a single non-semantic decision rule that maps all test inputs
to one class.

\textbf{Why B$\to$A partially succeeds.} Notri-Fact's uniform
article lengths \emph{force} the model to learn content-based semantic
features to discriminate between classes. These features -- lexical
choice, rhetorical style, sentence structure -- exhibit partial overlap
with semantic differences present in Dataset~A.

\subsubsection{Length Ablation: Experiment~5}

Truncating all Dataset~A articles to 50 words (primarily affecting
the fake class) yields F1~= 0.922 vs.\ 0.929 at full length -- a
marginal drop of only 0.007 points. This confirms that the length
confound inflates in-domain performance only marginally; the semantic
content within 50 words is also partially learnable. Crucially, there
was no significant performance drop after length capping, consistent
with the confound hypothesis: the shortcut is exploited but the
underlying semantic content within 50 words retains discriminative
signal. Given this negligible performance difference, a separate
figure for the length ablation is omitted; the result is discussed
here for completeness.

%%-- Confusion matrices: Urdu FND --%%

\begin{figure*}[!htb]
    \centering
    \begin{subfigure}[b]{0.495\textwidth}
        \includegraphics[width=\linewidth]{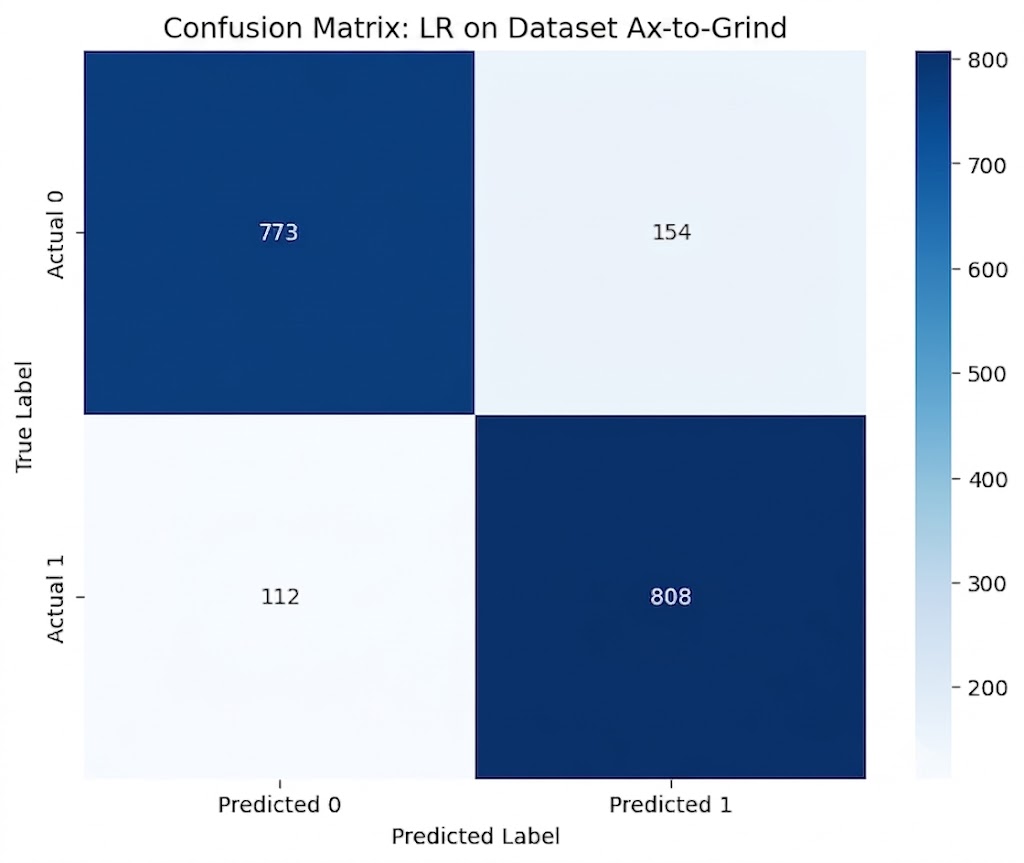}
        \caption{LR on Dataset A (Ax-to-Grind)}
    \end{subfigure}
    \hfill
    \begin{subfigure}[b]{0.495\textwidth}
        \includegraphics[width=\linewidth]{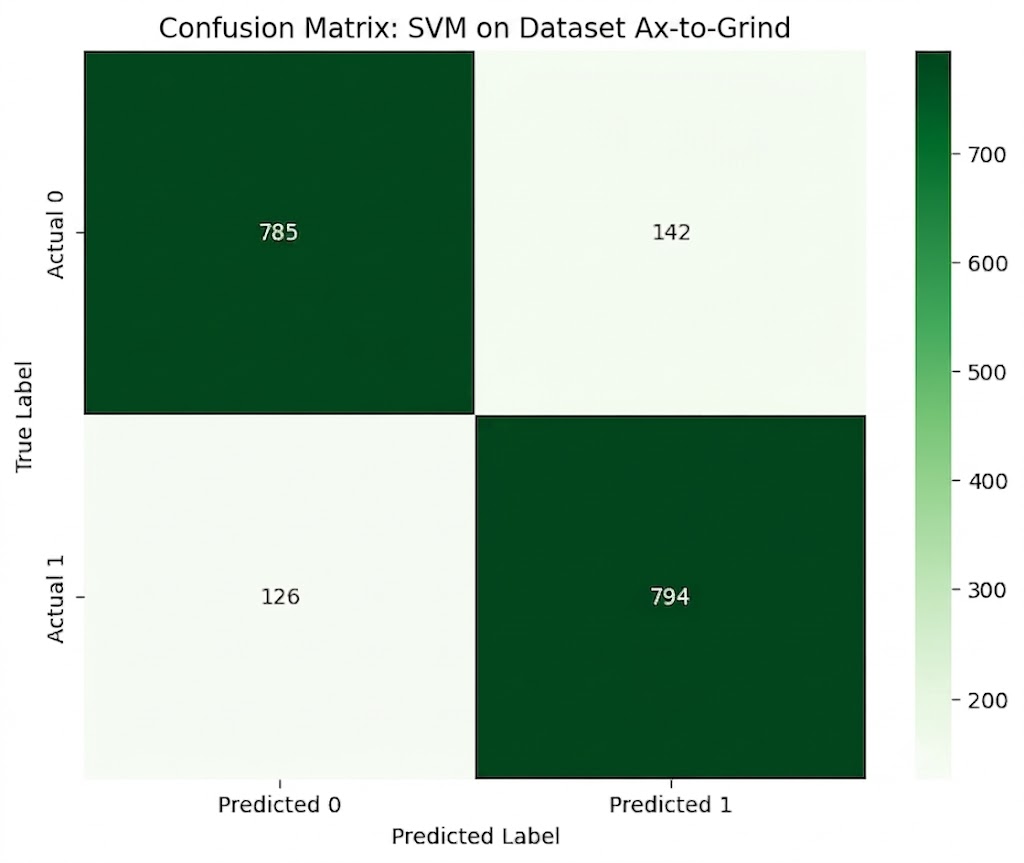}
        \caption{SVM on Dataset A (Ax-to-Grind)}
    \end{subfigure}
    \caption{Confusion matrices: TF-IDF classifiers on Ax-to-Grind (Dataset A).
    Both LR and SVM exhibit higher false positive rates for real news predicted as fake,
    consistent with partial exploitation of the length signal.}
    \label{fig:cm_urdu_a}
\end{figure*}

\begin{figure*}[!htb]
    \centering
    \begin{subfigure}[b]{0.495\textwidth}
        \includegraphics[width=\linewidth]{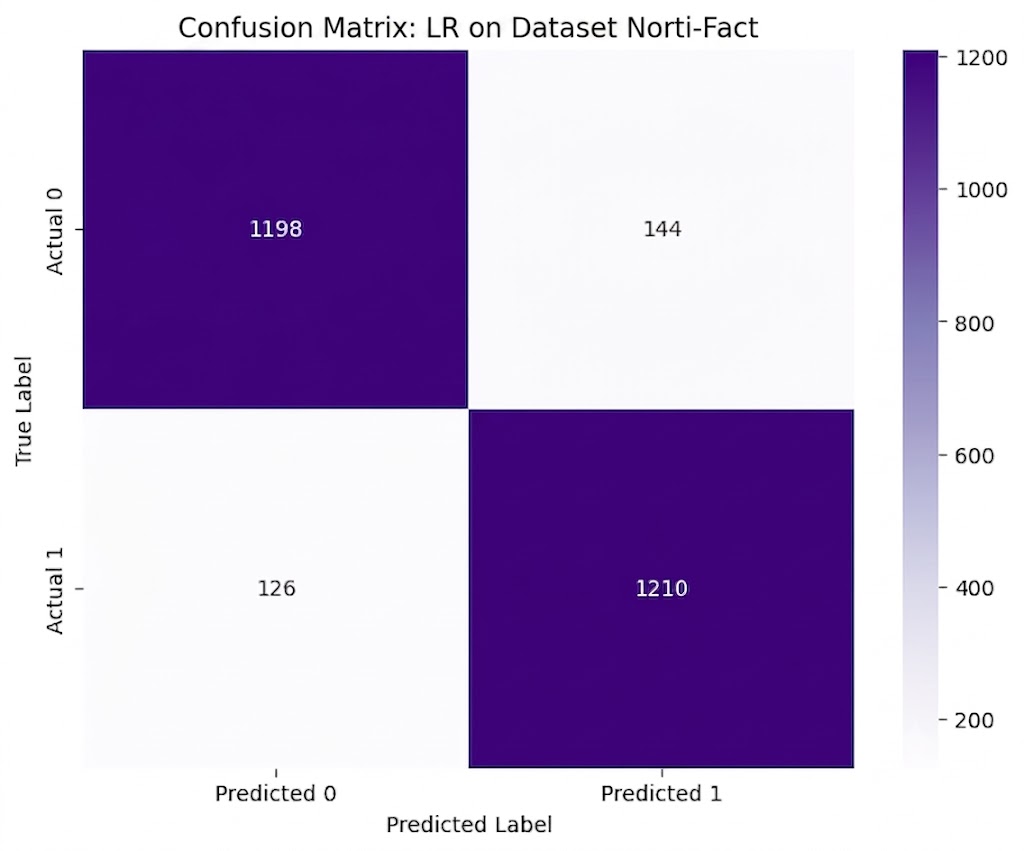}
        \caption{LR on Dataset B (Notri-Fact)}
    \end{subfigure}
    \hfill
    \begin{subfigure}[b]{0.495\textwidth}
        \includegraphics[width=\linewidth]{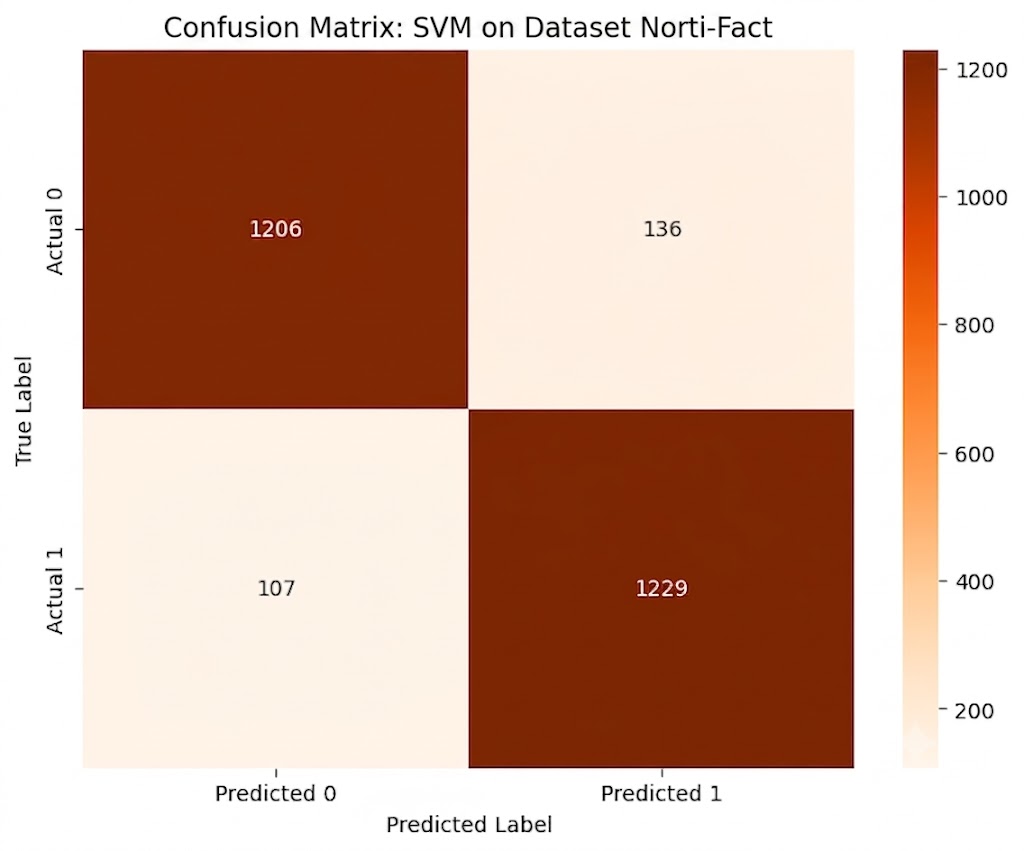}
        \caption{SVM on Dataset B (Notri-Fact)}
    \end{subfigure}
    \caption{Confusion matrices: TF-IDF classifiers on Notri-Fact (Dataset B).
    Near-perfect confusion matrices confirm that high in-domain accuracy is achievable
    without a length shortcut when classes are length-balanced.}
    \label{fig:cm_urdu_b}
\end{figure*}

\subsection{English FND Results}

Table~\ref{tab:main_results_en} presents complete results for
Experiments~6--9.

\begin{table*}[!htb]
\centering
\caption{Full Results: English Fake News Detection}
\label{tab:main_results_en}
\begin{tabular}{llcccc}
\toprule
\textbf{Exp} & \textbf{Model} & \textbf{F1} & \textbf{Acc} &
\textbf{Prec} & \textbf{Rec} \\
\midrule
\multicolumn{6}{l}{\textit{Dataset C -- WELFake (in-domain, 20\% test split)}} \\
6 & XLM-RoBERTa  & 0.986 & 0.988 & 0.986 & 0.986 \\
- & TF-IDF + LR  & 0.940 & 0.940 & 0.940 & 0.940 \\
- & TF-IDF + SVM & 0.950 & 0.950 & 0.950 & 0.950 \\
\midrule
\multicolumn{6}{l}{\textit{Dataset D -- ISOT (in-domain, 20\% test split)}} \\
7 & XLM-RoBERTa  & 1.000 & 1.000 & 1.000 & 1.000 \\
- & TF-IDF + LR  & 1.000 & 1.000 & 1.000 & 1.000 \\
- & TF-IDF + SVM & 1.000 & 1.000 & 1.000 & 1.000 \\
\midrule
\multicolumn{6}{l}{\textit{Cross-Domain Transfer (zero-shot, full dataset)}} \\
8 & XLM-RoBERTa (C$\to$D) & \textbf{0.006} & 0.500 & 0.250 & 0.500 \\
9 & XLM-RoBERTa (D$\to$C) & 0.251 & 0.501 & 0.251 & 0.501 \\
\bottomrule
\end{tabular}
\end{table*}

\subsubsection{In-Domain Performance}

XLM-RoBERTa achieves near-perfect in-domain performance: F1~= 0.986
on WELFake (Dataset~C) and F1~= 1.000 on ISOT (Dataset~D). TF-IDF
baselines are similarly strong: on WELFake, LR achieves F1~= 0.940
and SVM achieves F1~= 0.950. On ISOT, both LR and SVM achieve
F1~= 1.000 -- perfect in-domain classification. The perfect ISOT
in-domain scores suggest that the dataset contains strong
discriminative signals, possibly including stylistic features (all
real news from Reuters, fake from specific flagged sites) that make
in-domain classification trivially learnable even for TF-IDF models.

\subsubsection{Cross-Domain Transfer}

Both transfer directions collapse catastrophically, extending the
pattern found in Urdu FND to English. \textbf{Experiment~8
(C$\to$D, WELFake $\to$ ISOT):} F1~= 0.006, near-identical to
the Urdu A$\to$B collapse. The model trained on WELFake predicts
nearly all articles as one class when applied to ISOT.
\textbf{Experiment~9 (D$\to$C, ISOT $\to$ WELFake):} F1~= 0.251,
which while higher than Experiment~8, still represents catastrophic
failure -- far below both the in-domain baseline (F1~= 0.986) and
the Urdu B$\to$A result (F1~= 0.771).

A key factor is the extreme source specificity of ISOT: all real
articles originate exclusively from Reuters, while all fake articles
come from specific flagged misinformation sites. This creates
\textbf{publisher-level shortcuts} -- distinctive writing styles,
boilerplate, and phrasing patterns specific to Reuters and the flagged
sites. A model trained on ISOT learns to recognise Reuters as ``real''
rather than learning genuine semantic indicators of veracity. When
applied to WELFake, which aggregates from four different sources with
no Reuters dominance, the learned publisher shortcut provides no
signal. Referring to Fig.~\ref{fig:distEN_fake}--\ref{fig:distEN_isot_real},
WELFake articles span a wider length range with a higher upper tail
(up to 14,000 words) compared to ISOT (up to 8,000 words), and the
moderate WELFake length asymmetry (892 vs.\ 679 words) likely also
contributes to the C$\to$D collapse, analogously to the Urdu case.

%%-- Confusion matrices: English FND --%%

\begin{figure*}[!htb]
    \centering
    \begin{subfigure}[b]{0.495\textwidth}
        \includegraphics[width=\linewidth]{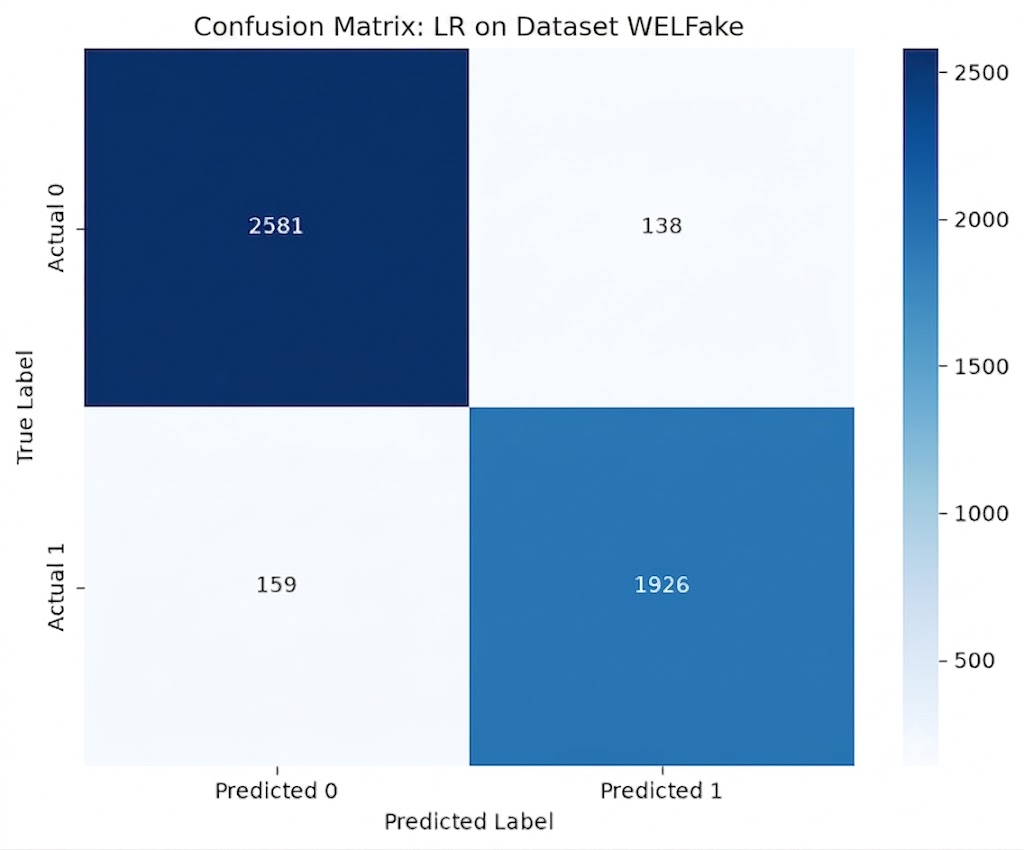}
        \caption{LR on Dataset C (WELFake)}
    \end{subfigure}
    \hfill
    \begin{subfigure}[b]{0.495\textwidth}
        \includegraphics[width=\linewidth]{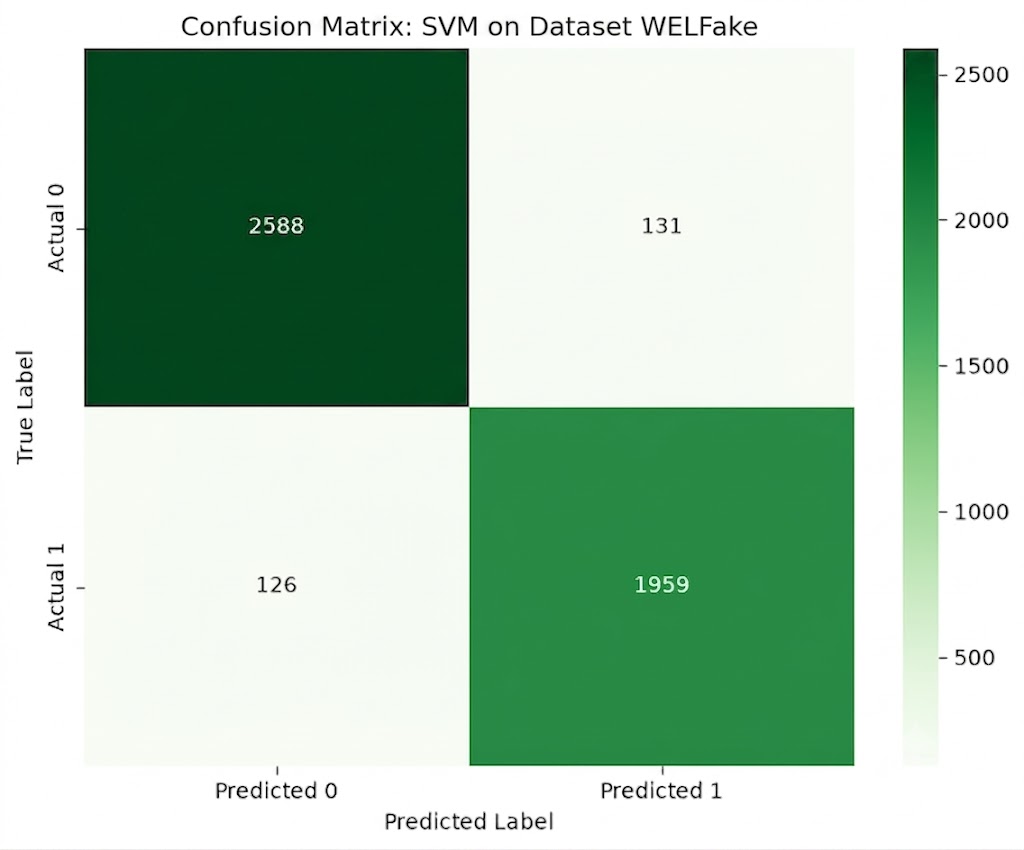}
        \caption{SVM on Dataset C (WELFake)}
    \end{subfigure}
    \caption{Confusion matrices: TF-IDF classifiers on WELFake (Dataset C).}
    \label{fig:cm_en_c}
\end{figure*}

\begin{figure*}[!htb]
    \centering
    \begin{subfigure}[b]{0.495\textwidth}
        \includegraphics[width=\linewidth]{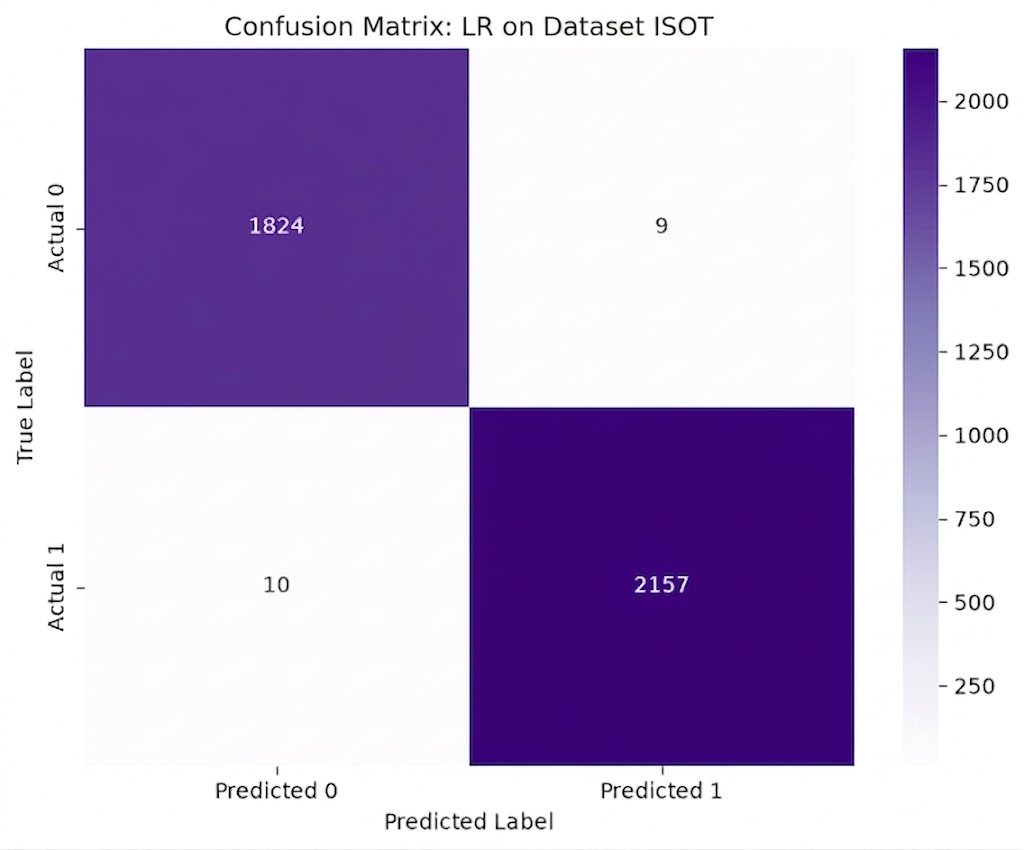}
        \caption{LR on Dataset D (ISOT)}
    \end{subfigure}
    \hfill
    \begin{subfigure}[b]{0.495\textwidth}
        \includegraphics[width=\linewidth]{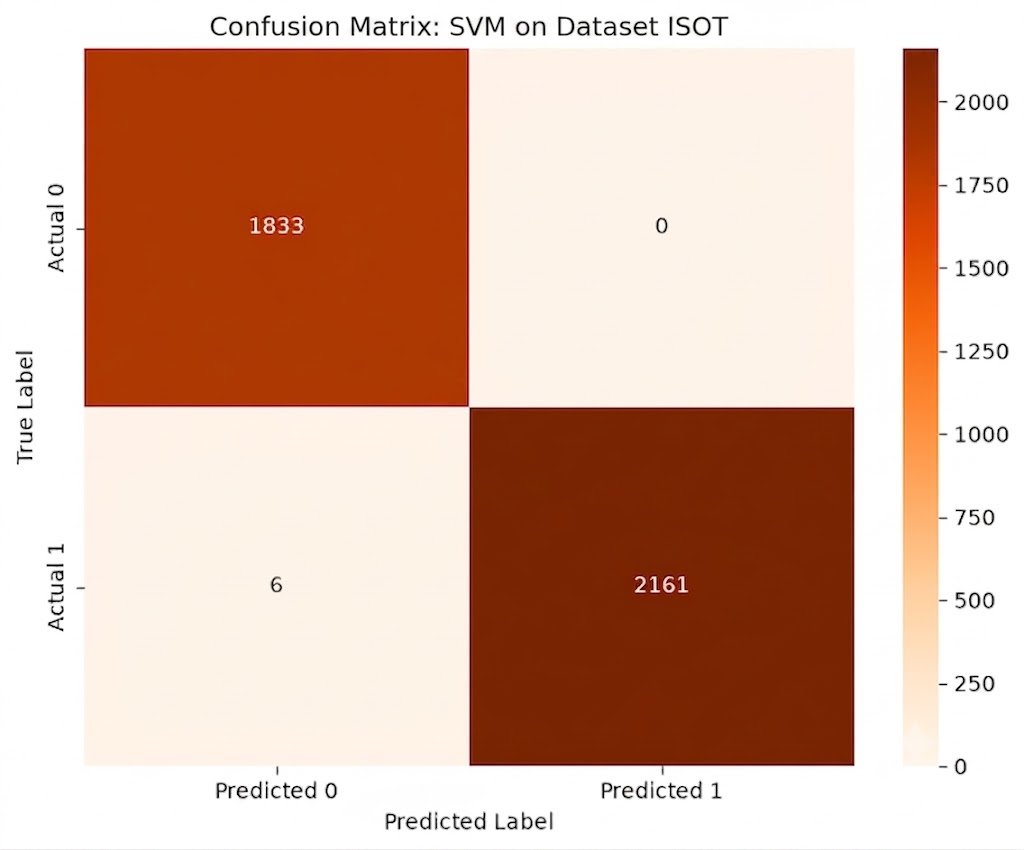}
        \caption{SVM on Dataset D (ISOT)}
    \end{subfigure}
    \caption{Confusion matrices: TF-IDF classifiers on ISOT (Dataset D).
    Near-perfect classification reflects strong source-level separability.}
    \label{fig:cm_en_d}
\end{figure*}

\subsection{Sarcasm Detection Results}

Table~\ref{tab:main_results_sarc} presents complete results for
Experiments~10--13.

\begin{table*}[!htb]
\centering
\caption{Full Results: Sarcasm Detection. $^{\star}$E$\to$F attempted multiple times; consistently collapsed to near-constant prediction (F1~$\approx$~0.004).}
\label{tab:main_results_sarc}
\begin{tabular}{llcccc}
\toprule
\textbf{Exp} & \textbf{Model} & \textbf{F1} & \textbf{Acc} &
\textbf{Prec} & \textbf{Rec} \\
\midrule
\multicolumn{6}{l}{\textit{Dataset E -- TweetEval Irony (in-domain, 20\% test split)}} \\
10 & XLM-RoBERTa  & 0.688 & 0.685 & 0.688 & 0.685 \\
-  & TF-IDF + LR  & 0.660 & 0.670 & 0.670 & 0.660 \\
-  & TF-IDF + SVM & 0.660 & 0.660 & 0.660 & 0.660 \\
\midrule
\multicolumn{6}{l}{\textit{Dataset F -- Sarcasm Corpus V2 (in-domain, 20\% test split)}} \\
11 & XLM-RoBERTa  & 0.788 & 0.787 & 0.789 & 0.788 \\
-  & TF-IDF + LR  & 0.730 & 0.730 & 0.730 & 0.730 \\
-  & TF-IDF + SVM & 0.720 & 0.720 & 0.720 & 0.720 \\
\midrule
\multicolumn{6}{l}{\textit{Cross-Domain Transfer (zero-shot, full dataset)}} \\
12 & XLM-RoBERTa (E$\to$F) & \textbf{0.004}$^{\star}$ & 0.500 & 0.250 & 0.500 \\
13 & XLM-RoBERTa (F$\to$E) & 0.642 & 0.640 & 0.643 & 0.642 \\
\bottomrule
\end{tabular}
\end{table*}

\subsubsection{In-Domain Performance}

XLM-RoBERTa achieves F1~= 0.688 on TweetEval Irony (Dataset~E) and
F1~= 0.788 on Sarcasm Corpus V2 (Dataset~F). These scores are
substantially lower than the fake news in-domain results, reflecting
the inherent difficulty of sarcasm and irony detection as a
classification task -- sarcasm relies on pragmatic, contextual cues
that are challenging to encode from text alone. TF-IDF baselines are
close to the transformer: LR achieves F1~= 0.660 on TweetEval and
F1~= 0.730 on Sarcasm Corpus V2, with SVM performing similarly.
The narrow transformer--TF-IDF gap in sarcasm detection (2.8--5.8
points) contrasts with the larger gaps seen in fake news tasks,
suggesting that surface lexical features carry proportionally more
discriminative signal for sarcasm.

\subsubsection{Cross-Domain Transfer -- Experiment~13 (F$\to$E)}

The reverse transfer direction (Sarcasm Corpus V2 $\to$ TweetEval)
achieves F1~= 0.642. While this represents a 14.6-point drop from
the in-domain F1~= 0.788 on Dataset~F, the model retains meaningful
discriminative ability on the target domain. This mirrors the B$\to$A
result in Urdu FND (F1~= 0.771 with a 18.3-point drop): when the
training corpus has relatively balanced length distributions, the
model is encouraged to learn semantic sarcasm markers that exhibit
partial transferability.

\subsubsection{Cross-Domain Transfer -- Experiment~12 (E$\to$F): Predicted Collapse}

Experiment~12 (TweetEval $\to$ Sarcasm Corpus V2) was attempted multiple
times across independent runs and consistently produced near-zero macro
F1 ($\approx$0.004), with the model predicting a single class for
virtually all Sarcasm Corpus V2 examples -- the same predicted-label
collapse pattern observed in Experiments~3 and~8. The collapse is
structurally predictable: TweetEval tweets average 13.7 words while
Sarcasm Corpus V2 forum posts average 48.7 words, a cross-corpus
length gap of $\approx$3.5$\times$ that closely mirrors the Urdu
Ax-to-Grind ratio (3.4$\times$). A model trained exclusively on
short tweet-length inputs encodes length-related signals as a
sarcasm proxy; encountering substantially longer forum texts, it
collapses to a constant prediction. The result is reported as
F1~= 0.004 (marked $\star$) and is treated as a collapse outcome
consistent with the other two domains rather than a missing value.

%%-- Confusion matrices: Sarcasm --%%

\begin{figure*}[!htb]
    \centering
    \begin{subfigure}[b]{0.495\textwidth}
        \includegraphics[width=\linewidth]{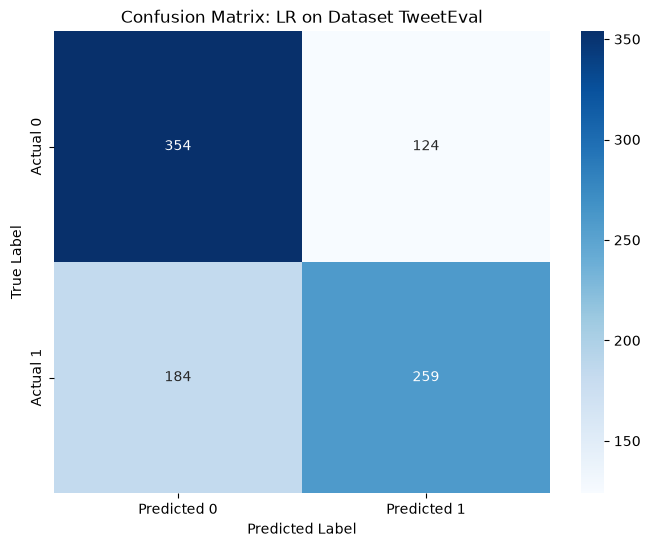}
        \caption{LR on Dataset E (TweetEval Irony)}
    \end{subfigure}
    \hfill
    \begin{subfigure}[b]{0.495\textwidth}
        \includegraphics[width=\linewidth]{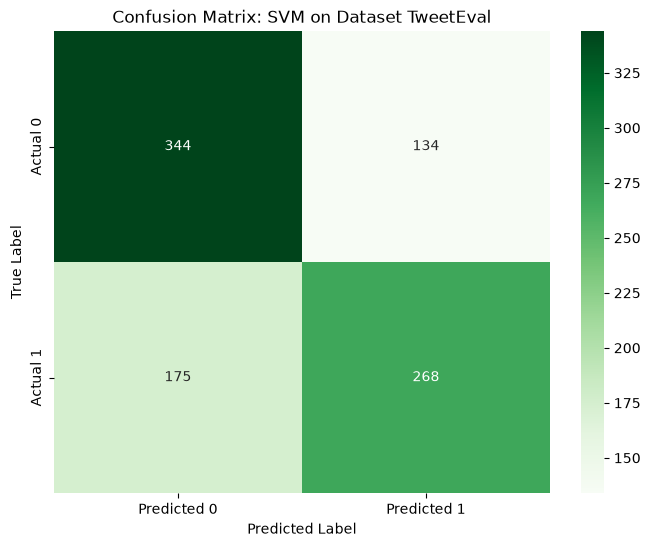}
        \caption{SVM on Dataset E (TweetEval Irony)}
    \end{subfigure}
    \caption{Confusion matrices: TF-IDF classifiers on TweetEval Irony (Dataset E).}
    \label{fig:cm_sarc_e}
\end{figure*}

\begin{figure*}[!htb]
    \centering
    \begin{subfigure}[b]{0.495\textwidth}
        \includegraphics[width=\linewidth]{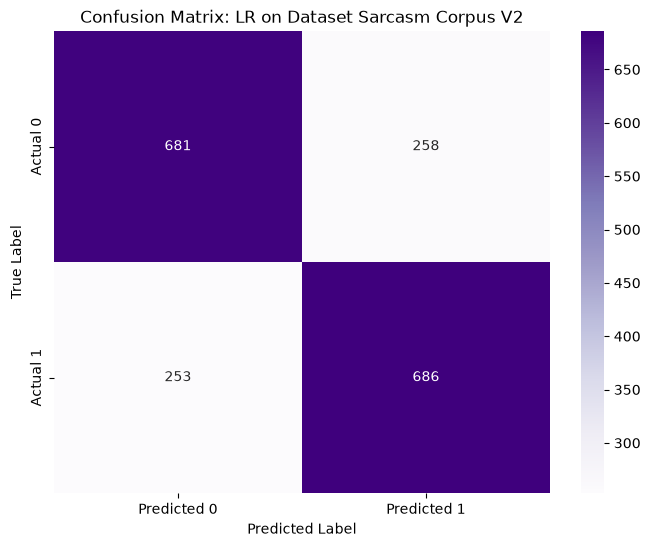}
        \caption{LR on Dataset F (Sarcasm Corpus V2)}
    \end{subfigure}
    \hfill
    \begin{subfigure}[b]{0.495\textwidth}
        \includegraphics[width=\linewidth]{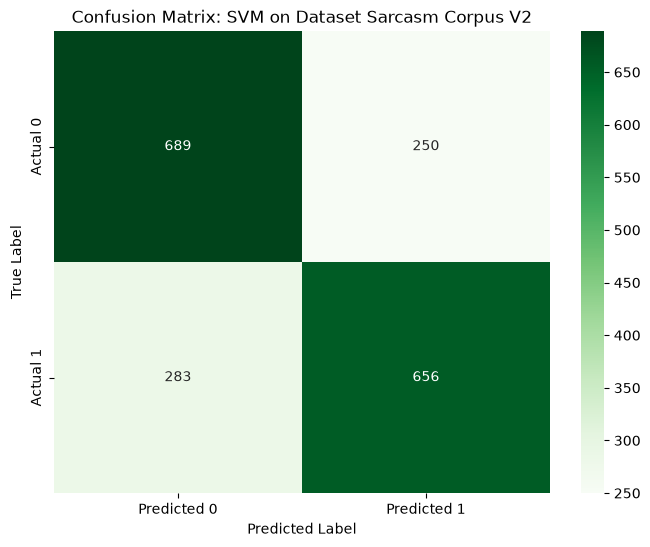}
        \caption{SVM on Dataset F (Sarcasm Corpus V2)}
    \end{subfigure}
    \caption{Confusion matrices: TF-IDF classifiers on Sarcasm Corpus V2 (Dataset F).}
    \label{fig:cm_sarc_f}
\end{figure*}

\subsection{Summary: All Results Ranked by F1-Score}

Table~\ref{tab:summary} consolidates all 13 experiments and TF-IDF
baselines.

\begin{table*}[!htb]
\centering
\caption{All Results Ranked by F1-Score.}
\label{tab:summary}
\setlength{\tabcolsep}{4pt}
\begin{tabular}{llp{5cm}cc}
\toprule
\textbf{Domain} & \textbf{Model} & \textbf{Condition} & \textbf{F1} & \textbf{Acc} \\
\midrule
Urdu   & XLM-RoBERTa & In-domain B         & 0.954 & 0.953 \\
Urdu   & XLM-RoBERTa & In-domain A (full)  & 0.929 & 0.929 \\
Urdu   & XLM-RoBERTa & In-domain A (50w)   & 0.922 & 0.922 \\
Urdu   & TF-IDF+SVM  & In-domain B         & 0.910 & 0.910 \\
Urdu   & TF-IDF+LR   & In-domain B         & 0.900 & 0.900 \\
Urdu   & TF-IDF+LR   & In-domain A         & 0.860 & 0.860 \\
Urdu   & TF-IDF+SVM  & In-domain A         & 0.855 & 0.855 \\
Urdu   & XLM-RoBERTa & Cross B$\to$A       & 0.771 & 0.773 \\
Urdu   & XLM-RoBERTa & Cross A$\to$B       & 0.005 & 0.501 \\
\midrule
EN FND & XLM-RoBERTa & In-domain D (ISOT)  & 1.000 & 1.000 \\
EN FND & TF-IDF+LR   & In-domain D (ISOT)  & 1.000 & 1.000 \\
EN FND & TF-IDF+SVM  & In-domain D (ISOT)  & 1.000 & 1.000 \\
EN FND & XLM-RoBERTa & In-domain C (WELFake) & 0.986 & 0.988 \\
EN FND & TF-IDF+SVM  & In-domain C (WELFake) & 0.950 & 0.950 \\
EN FND & TF-IDF+LR   & In-domain C (WELFake) & 0.940 & 0.940 \\
EN FND & XLM-RoBERTa & Cross D$\to$C       & 0.251 & 0.501 \\
EN FND & XLM-RoBERTa & Cross C$\to$D       & 0.006 & 0.500 \\
\midrule
Sarc.  & XLM-RoBERTa & In-domain F (SarcV2)  & 0.788 & 0.787 \\
Sarc.  & TF-IDF+LR   & In-domain F (SarcV2)  & 0.730 & 0.730 \\
Sarc.  & TF-IDF+SVM  & In-domain F (SarcV2)  & 0.720 & 0.720 \\
Sarc.  & XLM-RoBERTa & In-domain E (TweetEval) & 0.688 & 0.685 \\
Sarc.  & TF-IDF+LR   & In-domain E (TweetEval) & 0.660 & 0.670 \\
Sarc.  & TF-IDF+SVM  & In-domain E (TweetEval) & 0.660 & 0.660 \\
Sarc.  & XLM-RoBERTa & Cross F$\to$E       & 0.642 & 0.640 \\
Sarc.  & XLM-RoBERTa & Cross E$\to$F       & 0.004$^{\star}$ & 0.500 \\
\bottomrule
\end{tabular}
\end{table*}

\subsection{Confusion Matrix Analysis}

Figures~\ref{fig:cm_urdu_a}--\ref{fig:cm_sarc_f}
show TF-IDF baseline confusion matrices for all six datasets.
On Dataset~A (Ax-to-Grind), both LR and SVM exhibit
higher false positive rates for real news (predicted as fake),
consistent with partial exploitation of the length signal. On
Dataset~B (Notri-Fact), the near-perfect confusion matrices confirm
that high in-domain accuracy is achievable without a length shortcut
when classes are length-balanced. On Dataset~D (ISOT), both
classifiers achieve near-perfect classification, reflecting the
extreme source-level separability of that corpus. The sarcasm
confusion matrices show the more symmetric error patterns
characteristic of tasks where shortcut-free classification is
genuinely harder.

\clearpage
%%====================================================================
\section{Discussion}
\label{sec:discussion}
%%====================================================================

\subsection{Implications for Dataset Construction}

The length confound in Ax-to-Grind is not a data quality failure but
an artefact of collection methodology: authentic breaking news is
brief wire copy; fabricated articles from misinformation sites are
elaborated opinion pieces. This real-world asymmetry is acknowledged
in English FND literature~\cite{zhou2020survey} but has not previously
been analysed for its effect on cross-dataset generalisation in Urdu.

High in-domain performance on Ax-to-Grind -- F1~= 0.924 as reported
by Harris et al.~\cite{harris2024axtogrind} and 0.929 replicated here
-- may significantly overstate genuine discriminative capability.
Any model exploiting document length will score well in-domain but
collapse when the target dataset has a different length distribution.

We recommend for future FND and sarcasm dataset construction:
\begin{enumerate}
    \item Report \textbf{per-class length statistics} as mandatory
    metadata for every released corpus.
    \item \textbf{Audit for length confounds} and apply
    length-stratified sampling where needed.
    \item Make \textbf{cross-dataset transfer experiments} part of
    the standard evaluation protocol alongside single-dataset splits.
    \item Report the \textbf{cross-dataset generalisation gap}
    (in-domain F1 minus zero-shot cross-domain F1) as a standard
    diagnostic metric.
\end{enumerate}

\subsection{Generalisation Across Domains}

Our results demonstrate that transfer collapse is not a phenomenon
unique to Urdu FND. In English FND, both transfer directions fail:
Experiment~8 (WELFake $\to$ ISOT) collapses to F1~= 0.006,
matching the severity of the Urdu A$\to$B collapse, while
Experiment~9 (ISOT $\to$ WELFake) yields F1~= 0.251 -- better than
random but far below any useful threshold. The English FND failures
are driven by both the moderate WELFake length asymmetry
(892 vs.\ 679 words) and the extreme publisher-level shortcut in ISOT
(all real news from Reuters). In sarcasm detection, the partial
transferability observed in Experiment~13 (F$\to$E, F1~= 0.642)
suggests that when a model is trained on a length-balanced corpus
(Sarcasm Corpus V2, where both classes have similar distributions),
some semantic sarcasm markers do transfer to a structurally different
platform. The E$\to$F direction could not be completed due to a
technical error, but the cross-corpus length gap ($\approx$3.5$\times$)
strongly predicts a collapse analogous to the other two domains.

The domain pair E/F (TweetEval vs.\ Sarcasm Corpus V2) presents a
particularly instructive case: the two corpora differ in both
\textbf{platform} (Twitter vs.\ online forum), \textbf{length}
(short tweets vs.\ longer debate posts), and \textbf{sarcasm style}
(social-media irony markers vs.\ argumentative sarcasm). Each of
these dimensions constitutes an independent source of distributional
shift. Unlike the Urdu case, where the confound is concentrated in a
single dimension (article length), sarcasm detection transfer failure
likely reflects a combination of length mismatch and semantic register
mismatch, making diagnosis more complex and motivating domain-adaptive
approaches beyond length debiasing alone.

\subsection{Implications for Model Evaluation}

The standard single-dataset 80/20 evaluation cannot detect shortcut
learning: the shortcut is equally present in both splits. A more
rigorous protocol requires: (1) in-domain held-out performance,
(2) zero-shot cross-dataset performance on at least one additional
corpus, and (3) length-controlled ablation confirming that accuracy
is not substantially driven by the length confound.

\subsection{Relation to Shortcut Learning Literature}

Our findings align with and extend the shortcut learning literature to
fake news and sarcasm detection. Geirhos et
al.~\cite{geirhos2020shortcut} characterise shortcut learning as a
general failure mode of empirical risk minimisation: models learn the
easiest predictive rule available in training data, which may not
generalise. Wang and Culotta~\cite{wang2021spurious} demonstrate this
in text classification, showing that spurious surface correlations
cause performance to degrade under distributional shift. Our results
across three languages and domains extend these findings: the
length-based shortcut mechanism documented in NLP by Gururangan et
al.~\cite{gururangan2018annotation} and McCoy et
al.~\cite{mccoy2019right} in NLI is equally operative in fake news
detection and sarcasm detection, including in the Urdu low-resource
setting where such confounds have not previously been reported. The
proposed diagnostic methodology -- bidirectional transfer combined
with prediction collapse analysis -- is a practical tool for future
confound detection in any low-resource setting.

\subsection{Limitations}

Several limitations bound the scope of this study. First, we evaluate
only \texttt{xlm-roberta-base}; larger models (XLM-RoBERTa-large,
mDeBERTa) or Urdu-specific pre-trained models may exhibit different
shortcut behaviour. Second, we do not investigate domain adaptation
techniques (continued pre-training, adapter layers, adversarial length
debiasing) that could close cross-domain gaps. Third, the Urdu length
ablation (Exp~5) reveals a marginal 0.007-point F1 drop, suggesting
that while length is exploited as a shortcut, the semantic content
within 50 words also carries discriminative signal; full causal
isolation warrants more aggressive controls. Fourth, Notri-Fact's
annotation methodology is less documented than Ax-to-Grind's, leaving
uncertainty about label quality. Fifth, Experiment~12 (E$\to$F)
could not be completed due to a technical schema error, leaving one
cross-domain result absent from the sarcasm analysis.

%%====================================================================
\section{Conclusion}
\label{sec:conclusion}
%%====================================================================

We presented the first cross-dataset generalisation study for Urdu
fake news detection and extended the investigation to English FND and
sarcasm detection. Fine-tuning XLM-RoBERTa on Ax-to-Grind achieves
F1~= 0.929 in-domain but collapses to F1~= 0.005 on zero-shot
transfer to Notri-Fact, with 99.7\% of balanced test articles
predicted fake. The reverse direction achieves F1~= 0.771. Through
class-conditional length analysis and predicted label distribution
inspection, we identify a severe \textbf{length confound} in
Ax-to-Grind (fake articles 3.4$\times$ longer) as the root cause.
The length ablation (50-word cap, Exp~5) confirms that the confound
inflates but does not solely drive in-domain performance (F1 drop:
0.007).

Extension to English FND (WELFake $\leftrightarrow$ ISOT) and
sarcasm detection (TweetEval $\leftrightarrow$ Sarcasm Corpus V2)
reveals that transfer collapse is a cross-lingual, cross-domain
challenge in binary text classification. English FND exhibits
particularly severe collapse in both directions (C$\to$D: F1~= 0.006;
D$\to$C: F1~= 0.251), driven by publisher-level shortcuts in ISOT and
a moderate length asymmetry in WELFake. Sarcasm detection shows
partial transferability in the F$\to$E direction (F1~= 0.642),
suggesting that length-balanced training encourages learning of more
transferable semantic features. These findings confirm that shortcut
learning from dataset-specific distributional artefacts is a
\textbf{universal challenge} in binary text classification, not a
peculiarity of Urdu or low-resource settings.

These findings underscore that in-domain performance on a single
benchmark is insufficient to claim generalisation. We propose
cross-dataset evaluation as a mandatory complement to in-domain
benchmarking and recommend that future FND and sarcasm datasets
report per-class length statistics and conduct transfer experiments
as standard practice.

Future work will investigate \textbf{length-invariant training
strategies} (length normalisation, curriculum learning, adversarial
debiasing), domain-adaptive fine-tuning to close cross-domain gaps,
and extension to additional corpora in other low-resource languages.

\section*{Reproducibility}

All experiments used Python with Hugging Face
\texttt{transformers}~v5.0~\cite{wolf2020transformers} and
\texttt{datasets}~v4.0, scikit-learn~1.6.1, PyTorch~2.10.0, and
CUDA~12.8 on an NVIDIA T4 GPU via Google Colab. Seed~42 throughout.
Dataset~A: \url{https://github.com/Sheetal83/Ax-to-Grind-Urdu-Dataset}.
Dataset~B: \url{https://www.kaggle.com/datasets/tridata/notri-fact-real-and-unreal-urdu-news}.
Dataset~C (WELFake): available on Kaggle as \texttt{WELFake\_Dataset.csv}.
Dataset~D (ISOT): \texttt{True.csv} and \texttt{Fake.csv} available on Kaggle.
Dataset~E: \url{https://huggingface.co/datasets/cardiffnlp/tweet_eval} (irony split).
Dataset~F: \url{https://nlds.soe.ucsc.edu/sarcasm2/}.
Notebooks and model checkpoints released upon acceptance.

%%====================================================================
\bibliographystyle{IEEEtran}

\end{document}